\documentclass[sigconf,screen,authorversion,nonacm]{acmart}

\usepackage{array}
\usepackage{stfloats}
\usepackage{url}
\usepackage{verbatim}
\usepackage{bbding}
\usepackage{graphicx}
\usepackage{textcomp}
\usepackage{xcolor}
\usepackage{multirow}
\usepackage{booktabs}
\usepackage{makecell}
\usepackage{mathrsfs}
\usepackage[ruled]{algorithm2e}
\newcommand{\eg}{\textit{e.g.}}
\newcommand{\ie}{\textit{i.e.}}
\newcommand{\our}{UniGAP\xspace}
\newcommand{\ours}{UniGAP\xspace}
\newtheorem{definition}{Definition}
\newtheorem{theorem}{Theorem}[section]
\newtheorem{lemma}[theorem]{Lemma}
\newtheorem{proposition}[theorem]{Proposition}

\newsavebox\CBox

\definecolor{dark2green}{rgb}{0.1, 0.65, 0.3}
\definecolor{dark2orange}{rgb}{0.9, 0.4, 0.}
\definecolor{dark2purple}{rgb}{0.4, 0.4, 0.8}

\DeclareMathOperator*{\argmin}{arg\,min}

\usepackage{dsfont}

\usepackage{hyperref}
\usepackage{tablefootnote}
\usepackage{colortbl}
\definecolor{LightCyan}{rgb}{0.88,1,1}
\definecolor{Gray}{gray}{0.92}

\AtBeginDocument{%
  }

\setcopyright{acmlicensed}
\copyrightyear{2018}
\acmYear{2018}
\acmDOI{XXXXXXX.XXXXXXX}

\acmConference[Conference acronym 'XX]{Make sure to enter the correct
  conference title from your rights confirmation emai}{June 03--05,
  2018}{Woodstock, NY}

\acmISBN{978-1-4503-XXXX-X/18/06}

\begin{document}

\title{UniGAP: A Universal and Adaptive Graph Upsampling Approach to Mitigate Over-Smoothing in Node Classification Tasks}

\author{Xiaotang Wang}
\affiliation{%
  \institution{The Hong Kong University of Science and Technology (Guangzhou)}
  \city{Guangzhou}
  \country{China}
  }
\email{xwang285@connect.hkust-gz.edu.cn}

\author{Yun Zhu}
\authornote{Corresponding author.}
\affiliation{%
  \institution{Shanghai Artificial Intelligence Laboratory}
  \city{Shanghai}
  \country{China}
  }
\email{zhuyun@pjlab.org.cn}

\author{Haizhou Shi}
\affiliation{%
 \institution{Rutgers University}
 \city{New Brunswick}
 \state{New Jersey}
 \country{US}
 }
 \email{haizhou.shi@rutgers.edu}

\author{Yongchao Liu}
\affiliation{%
  \institution{Ant Group}
  \city{Hangzhou}
  \country{China}
}
\email{yongchao.ly@antgroup.com}

\author{Yongqi Zhang}
\authornote{Principle Corresponding author.}
\affiliation{%
  \institution{The Hong Kong University of Science and Technology (Guangzhou)}
  \city{Guangzhou}
  \country{China}
}
\email{yongqizhang@hkust-gz.edu.cn}

\renewcommand{\shortauthors}{Wang et al.}

\begin{abstract}
In the graph domain, deep graph networks based on Message Passing Neural Networks~(MPNNs) or Graph Transformers often cause over-smoothing of node features, limiting their expressive capacity. 
Many upsampling techniques involving node and edge manipulation have been proposed to mitigate this issue. However, these methods are often heuristic, resulting in extensive manual labor and suboptimal performance and lacking a universal integration strategy.
In this study, we introduce UniGAP, a \underline{uni}versal and adaptive \underline{g}r\underline{a}ph u\underline{p}sampling framework
to mitigate over-smoothing in node classification tasks. Specifically, we design an adaptive graph upsampler based on condensed trajectory features, serving as a plug-in component for existing GNNs to mitigate the over-smoothing problem and enhance performance. Moreover, UniGAP serves as a representation-based and fully differentiable framework to inspire further exploration of graph upsampling methods.
Through extensive experiments, UniGAP demonstrates significant improvements over heuristic data augmentation methods in various datasets and metrics. We analyze how graph structure evolves with UniGAP, identifying key bottlenecks where over-smoothing occurs, and providing insights into how UniGAP addresses this issue. Lastly, we show the potential of combining UniGAP with large language models (LLMs) to further improve downstream performance.
Our code is available at: \href{https://anonymous.4open.science/r/UniGAP}
{\texttt{https://anonymous.4open.science/r/UniGAP}}.
\end{abstract}

\maketitle

\input\section{Introduction}
\sloppy
Graph neural networks (GNNs)~\citep{gcn,gat,graphsage} are powerful tools for handling graph-structured data, demonstrating significant success in recent years~\cite{berg2017graph,Ying_2018,zheng2019gman,Guo2019AttentionBS,gaudelet2021utilising}.
In contrast to computer vision and natural language processing, where deep models often lead to significant improvements~\citep{he2015deep}, in the graph domain, increasing the number of model layers can cause node features to converge exponentially towards the same vector, a phenomenon known as ``over-smoothing''~\cite{rusch2023survey,theoreticoversmoothing,adaedge}. This convergence diminishes the model's expressive power and adversely affects its performance.

Previous research has explored a variety of strategies to mitigate the over-smoothing problem, including manipulating graph topology~\citep{adaedge,rong2020dropedge,graphcl}, refining model structure~\citep{kipf2017semisupervised,jknet}, and employing dynamical learning~\citep{adgcn,dgsln}. 
Recently, graph upsampling, a lightweight graph topology manipulation strategy, has proven effective in complementing and enhancing other techniques.
AdaEdge~\citep{adaedge} relies on prior knowledge encoded in a heuristic adaptation rule—namely, increasing intra-class connectivity and decreasing inter-class connectivity—while HalfHop~\citep{azabou2023halfhop} depends on a predefined random insertion strategy. Such heuristic designs result in suboptimal performance and lack of interpretability, making it difficult to understand their effectiveness or rationale.

To address the aforementioned challenges, we introduce \our, a \underline{uni}versal and adaptive \underline{g}r\underline{a}ph u\underline{p}sampling methodology. Specifically, different from existing methods that employ a heuristic upsampling strategy, \our adaptively calculates the probability of inserting an intermediate node for each pair of adjacent nodes in the original graph, generating a new upsampled graph through a differentiable sampling technique. Subsequently, \our iteratively refines these sampling probabilities based on evaluation metrics (e.g., accuracy or smoothness) of the upsampled graph in downstream tasks. This iterative process strives to yield an optimally upsampled graph suitable for downstream applications.
Additionally, \our serves as a fully differentiable representation-based framework to inspire further exploration of graph upsampling methods, in which its components are all modular and can be replaced in different ways.
Furthermore, \our enhances the interpretability of the over-smoothing issue within graph structure learning. By examining the augmented graph that achieves optimal performance, we can precisely identify where intermediate nodes are inserted. This analysis provides valuable insights into the root causes of the over-smoothing problem. 
Lastly, we demonstrate the potential of integrating LLMs with \our to further improve downstream performance.

Our contributions can be summarized as follows:
\begin{itemize}
    \item We propose \our, a universal and adaptive graph upsampling approach to mitigate over-smoothing in node classification and enhance downstream performance as a plug-in. Additionally, \our serves as a fully differentiable trajectory-based framework to inspire further exploration of graph upsampling methods.
    \item \our identifies the critical bottleneck where the over-smoothing problem occurs, offering interpretability to this issue from a graph structure learning perspective, supported by our provided visualizations.
    \item Through extensive experiments, \our shows significant improvements over heuristic data augmentation methods and integrates seamlessly with GNNs to enhance performance and address over-smoothing. Additionally, it can also be combined with LLMs for further performance boosts.
\end{itemize}

\section{Related Work}
\subsection{Over-Smoothing Problem in GNNs}\label{app:oversmooth}
Over-smoothing represents a prevalent research issue within the realm of graphs, garnering significant attention recently. This phenomenon substantially impedes the expressiveness of GNNs as the number of propagation layers increases, resulting in node features becoming progressively more indistinguishable. 
To assess the extent of over-smoothing, a variety of quantitative metrics have been established, including Mean Average Distance (MAD)~\citep{chen2019measuring}, Euclidean Distance~\citep{euclidean_distance}, Dirichlet Energy~\citep{rusch2023survey}, and others. 

Previous research~\citep{cai2020note, reviewsm, rusch2023survey} has explored various strategies to mitigate the over-smoothing problem. These solutions can be broadly categorized into three main approaches: manipulating graph topology, refining the model structure, and dynamic learning.
Firstly, manipulating graph topology involves modifying the original graph structure using dropout techniques (\eg, DropEdge~\citep{rong2020dropedge}, DropNode~\citep{feng2021graph}, DropMessage~\citep{Fang_2023}), graph upsampling~\citep{azabou2023halfhop, adaedge}, or others~\citep{han2022gmixup}. These methods aim to reduce message passing or enhance graph structure to alleviate over-smoothing.
Secondly, refining the model structure includes modifying message passing mechanisms~\citep{gprgnn} or utilizing some model tricks like residual connections~\citep{gcnii} and multi-hop information~\citep{jknet} to mitigate over-smoothing.
Lastly, dynamic learning strategies dynamically adjust the learning process of the model. For instance, DAGNN~\citep{dagnn} calibrates a retention score to preserve representations in each layer, DGSLN~\citep{dgsln} proposes a graph generation module to optimize the graph structure, and ADGCN~\citep{adgcn} employs attention for selective cross-layer information aggregation.

In this study, our focus is on graph upsampling due to its simplicity and effectiveness. To the best of our knowledge, previous studies on graph upsampling have relied on heuristic strategies that involve significant human effort and lack a universal framework for integration. Our work aims to address these challenges by introducing a universal and adaptive method.

\subsection{Graph Augmentations}\label{app:aug}
Graphs, as non-Euclidean data, primarily consist of entities and the relations between them. Therefore, augmentation techniques for graph data are inherently more complex compared to other domains such as computer vision and natural language processing. Graph augmentations aim to enhance model generalization, improve performance, and mitigate over-smoothing. These augmentations can be categorized into three main types: node-level, edge-level, and graph-level augmentations.
Node-level augmentations focus on individual nodes. For example, DropNode~\citep{graphcl} randomly sets some nodes as zero vectors, FeatureMask~\citep{featuremask} randomly masks partial features of each node, and FLAG~\citep{flag} adds a learnable perturbation to each node.
Edge-level augmentations manipulate edges within the graph. For instance, DropEdge~\citep{rong2020dropedge} and EdgeMask~\citep{graphcl} randomly remove existing edges based on specified probabilities. 
AdaEdge~\citep{adaedge} relies on a heuristic approach by adjusting intra-class and inter-class connections based on model classification outcomes. 
GAug~\citep{zhao2020dataaugmentationgraphneural} uses neural edge predictors to estimate edge probabilities.
Graph-level augmentations modify both nodes and edges simultaneously. GraphCL~\citep{graphcl} combines various augmentation techniques to modify the original graph. UGA~\citep{uga} proposes a unified graph augmentation framework for generalized contrastive learning that automatically selects and composes suitable augmentations for different graphs and tasks. Additionally, GCA~\citep{gca} proposes adaptive augmentation approaches based on metrics such as the connectivity of target nodes. HalfHop~\citep{azabou2023halfhop} suggests adding nodes to existing edges randomly to slow down message propagation and mitigate over-smoothing.

Unlike previous studies, our work aims to develop a universal framework capable of integrating existing graph augmentations, particularly focusing on graph upsampling. Additionally, this framework is designed to be adaptive and optimized without requiring extensive human intervention.

\subsection{Graph Structure Learning}\label{app:gsl}
Graph structure learning (GSL) has emerged as a complementary line of research to graph augmentation. While graph augmentations typically apply stochastic or rule-based transformations to generate alternative graph views for regularization or contrastive learning, GSL aims to explicitly learn an optimized graph structure jointly with node representations in an end-to-end manner. 
A large body of GSL methods focuses on learning or refining the adjacency matrix through edge rewiring, i.e., adding, deleting, or reweighting edges between existing nodes. Some approaches learn a latent graph from node features and optimize it together with GNN parameters for downstream prediction~\citep{idgl,lds,infomgf}. In parallel, a series of rewiring methods modifies the graph topology to alleviate over-smoothing or over-squashing, for example, by inserting long-range edges, sparsifying redundant connections, or reconfiguring edges based on curvature or edge-noise estimates~\citep{gpsrewire,diffwire}. These methods typically operate in the space of edge-level operations and keep the set of nodes fixed. 

Compared with these rewiring-based GSL approaches, \our takes a different perspective by performing graph upsampling through node insertion on existing edges. Instead of only deciding whether an edge should be added, removed, or reweighted, \our adaptively inserts intermediate nodes to effectively stretch selected edges, thus slowing down message passing along them. 

\section{UniGAP: A Universal and Adaptive Graph Upsampling Method}
\begin{figure*}[t] 
\centering 
\includegraphics[width=1\linewidth]{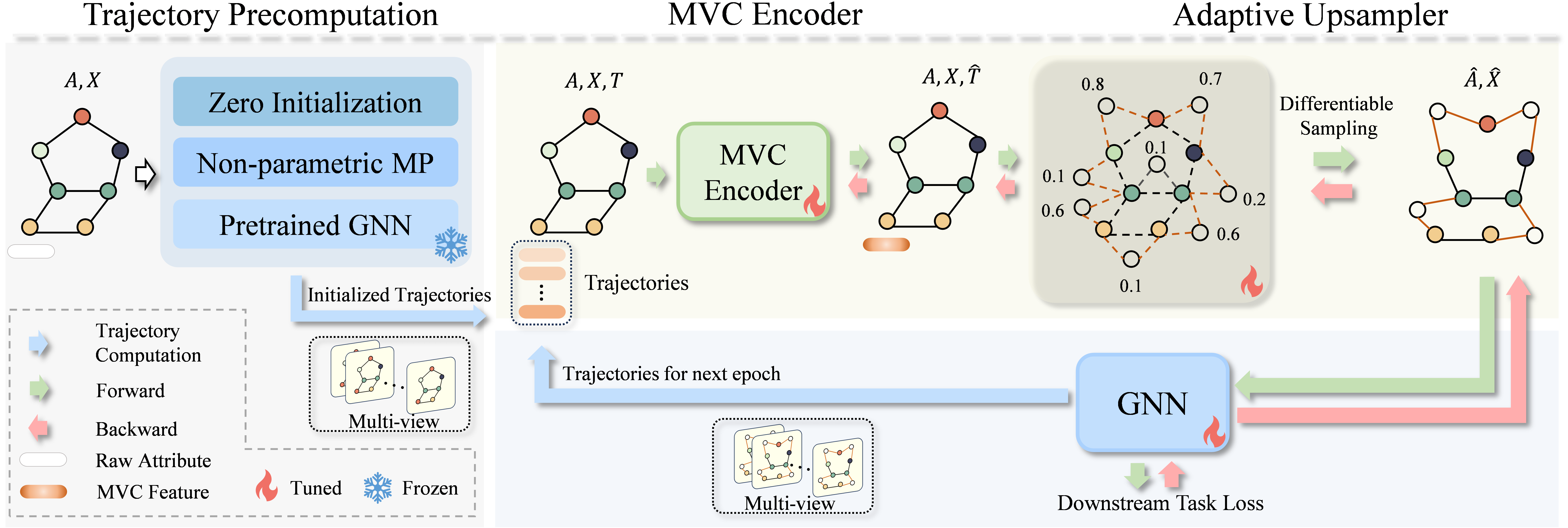} 
\caption{Overview of the \our framework. 
(1) Given an input graph, the \textbf{Trajectory Precomputation} module first computes layer-wise node trajectories that capture how node representations evolve with depth. 
(2) These trajectories are fed into the \textbf{Multi-View Condensation (MVC) Encoder}, which aggregates multi-hop information into a compact, per-node representation. 
(3) The \textbf{Adaptive Upsampler} then uses these condensed features to predict where to insert intermediate nodes on edges, and a differentiable sampling step produces an augmented graph with updated topology and node features. 
(4) Finally, the augmented graph is passed to a downstream GNN for tasks such as node classification, and the task loss is backpropagated to jointly update the MVC encoder, the upsampler, and the downstream model. 
After a warm-up epoch, the refined downstream GNN is reused to recompute trajectories, and the process is iterated until convergence.}
\label{framework} 
\end{figure*}

In this section, we introduce our proposed universal and adaptive graph upsampling method, \ie, UniGAP. Firstly, it computes the input graph's trajectories (multi-hop information) using our Trajectory Precomputation module~(Section~\ref{sec:trace}). 
These trajectories are then processed by a tunable Multi-View Condensation~(MVC) Encoder to extract condensed information~(Section~\ref{sec:mvc}). 
Next, the condensed features are used to compute node insertion probabilities via the adaptive upsampler module (Section~\ref{sec:upsampler}). 
Finally, the augmented graph is fed into the downstream model for specific tasks (Section~\ref{sec:training}). 
The pipeline and algorithmic description of UniGAP are shown in Figure~\ref{framework} and Algorithm~\ref{alg_our_method}, respectively. 

\textbf{Notations.}\quad
In this work, our focus is on node classification, which involves learning a graph predictor $g_\theta: \mathcal{G}\mapsto\mathcal{Y}$ that maps instance $G=(\mathcal{V},\mathcal{E},{A},{X}, {Y}) \in \mathcal{G}$ to label ${Y} \in \mathcal{Y}$, where \( \mathcal{V} \) and \( \mathcal{E} \) denote the sets of nodes and edges, respectively. \( {X} \in \mathbb{R}^{N \times d} \) represents the node attribute feature matrix with \( N \) nodes and \( d \) attributes. \( {A} \in \mathbb{R}^{N \times N} \) is the adjacency matrix.
Additionally, we denote the diagonal degree matrix as \( D \in \mathbb{R}^{N \times N} \), and the matrices with added self-loops as \( \tilde{A} \in \mathbb{R}^{N \times N} \) and \( \tilde{D} \in \mathbb{R}^{N \times N} \), respectively. The graph Laplacian with self-loops is defined as \( \tilde{L} = \tilde{D} - \tilde{A} \).
For normalization, the symmetrically normalized adjacency matrix is \( \hat{A} = \tilde{D}^{-\frac{1}{2}} \tilde{A} \tilde{D}^{-\frac{1}{2}} \), and the symmetrically normalized graph Laplacian is \( \hat{L} = \tilde{D}^{-\frac{1}{2}} \tilde{L} \tilde{D}^{-\frac{1}{2}} \). 
For clarity in notations, we will use \( A \) and \( L \) to denote the original graph, regardless of whether it is normalized or unnormalized. The augmented graph, which will be introduced later, will be represented by \( \hat{A} \) and \( \hat{L} \).

\subsection{Trajectory Computation}\label{sec:trace}
Considering the output of node features at each layer of the model ${H}^{(l)}$ implicitly reflects the gradual process of over-smoothing as the number of layers increases. We interpret these features as \textbf{trajectories} that model this process.
Concretely, for each node, the sequence of its representations across layers forms a trajectory that encodes how it is affected by over-smoothing as depth grows. Building on this intuitive assumption, our framework leverages these trajectories to diagnose where and when over-smoothing occurs and to design strategies that adaptively modify the graph structure to mitigate its effects.
Specifically, trajectory $T^{(l)}\in\mathbb{R}^{N\times d}$ refers to the hop information of each node, such as one-hop $H^{(1)}$, two-hop $H^{(2)}$, and higher-order features:
\begin{equation}
    \begin{aligned}
        {T}^{(l)}=\begin{cases}
                    \operatorname{Norm}({H}^{(l)})   , &\text{if} \ \operatorname{mod}(l,w)=0 \\
                    {H}^{(l)} , &\text{o.w.}               \\
                \end{cases}
    \end{aligned}
\end{equation}
where the representation of each hop is denoted as $H^{(l)}=f^{(l)}(H^{(l-1)})\in \mathbb{R}^{N\times d}$, $f^{(l)}(\cdot)$ means the $l$-th layer in the downstream model, and $\operatorname{Norm}({x}_{i})=\frac{{x}_{i}}{ \left| \left| {x}_{i} \right| \right|_{2}}$ signifies a channel-wise $L2$ normalization, which helps regulate the magnitude of features that may escalate exponentially with the number of iterations. $w$ is a hyperparameter that controls the frequency of normalization.

These trajectories are essential for the functionality of subsequent modules, which will be optimized iteratively. Let the trajectories be denoted as \( T=[{T}^{(1)}, {T}^{(2)}, \cdots, {T}^{(L)}] \in \mathbb{R}^{L \times N \times d} \), where \( L \) represents the number of layers, \( N \) denotes the number of nodes, and \( d \) signifies the dimensionality of the features. 

In the initial stage, a specific model is not available for calculating the trajectories of the original graph. Therefore, a trajectory precomputation module is needed to obtain initialized trajectories. Various strategies can be employed to achieve this goal. In this work, we explore three off-the-shelf strategies, considering their efficiency and simplicity: Zero Initialization, Non-parametric Message Passing, and Pretrained GNN. Next, we will discuss how to integrate these strategies into our methods: 

\textbf{1) Zero initialization.} We initialize the trajectories with all-zero vectors, resulting in zero multi-view condensed features during the first epoch of training. Consequently, the upsampling phase in this initial epoch does not introduce new edges or nodes, thus utilizing the original graph for downstream tasks. After finishing the first epoch, we store the hidden representations of each layer in downstream model as the trajectories for subsequent epochs. This zero initialization ensures that the \our module learns the graph structure tailored to each downstream task with minimal noise and optimal performance. The zero initialization can be formulated as
\begin{equation}
    {T}^{(l)}=\boldsymbol {0} \in \mathbb{R}^{N \times d}
\end{equation}

\textbf{2) Non-parametric Message Passing.} After removing the non-linear operator in graph neural networks, we can decouple the message passing ($AX$) and feature transformation ($XW$). However, in the initial stage, we have no access to the transformation weights $W$ unless pre-training methods are employed. Therefore, we adopt non-parameteric message passing to collect multi-view information, \ie, $AX, A^2X,\cdots A^LX$.
This approach avoids nonlinear transformations, preserving maximum original feature and pure graph topology information. The formulation is as follows:
\begin{equation}
    T^{(l)}=\mathcal{P}^{(l)}=AT^{(l-1)} \ \text{or} \ T^{(l)}=\mathcal{P}^{(l)}=LT^{(l-1)} 
\end{equation}
where $\mathcal{P}^{(l)}$ denotes the $l$-th layer in $\mathcal{P}$ and $T^{(0)}=G$.

\textbf{3) Pretrained GNN.} 
The first two methods do not introduce additional parameters, which may limit the expressivity of the information embedded in the raw features. Here, we introduce a more powerful method, though it may incur additional computational costs. We leverage the training data to train a GNN model by designing a pretext task such as unsupervised contrastive loss~\citep{graphcl,rosa,graphcontrol} or masked prediction~\citep{graphmae,zhu2023sgl} to obtain a generalized feature extractor. An additional advantage of this method is the ability to use pre-trained parameters to initialize the encoding parameters in the downstream model, facilitating faster convergence and reducing training time. In this way, we can kill two birds with one stone. The pretraining procedure can be formulated as follows:
\begin{equation}
\mathcal{P}=\argmin_\theta\mathcal{L}_{\text{pretext}}(\mathcal{P}_\theta(G))
\end{equation}
After the pre-training process, the pre-trained GNN model will be employed to compute trajectories:
\begin{equation}
    T^{(l)} = \mathcal{P}^{(l)}(T^{(l-1)})
\end{equation}

In Section~\ref{sec:ablation}, we experimentally explore various precomputation strategies to address the cold start problem in trajectory computation. Based on performance, we select Pretrained GNN as the precomputation approach for this work.

\subsection{Multi-View Condensation}\label{sec:mvc}
After obtaining trajectories (multi-view information), we aim to adaptively condense this multi-view information into compact features that reflect the node's propensity to over-smoothing, while considering efficiency and adaptivity. To achieve this goal, we design a tunable multi-view condensation (MVC) encoder \( \mathcal{F}_{\omega}(\cdot): \mathbb{R}^{L \times N \times d} \mapsto \mathbb{R}^{N \times d} \) that can adaptively condense trajectories \( T \in \mathbb{R}^{L \times N \times d} \) into condensed trajectory features \( \hat{T} \in \mathbb{R}^{N \times d} \). This reduces computational cost and adaptively adjusts the weights of information at each hop. In this work, we instantiate the MVC encoder using two strategies: Trajectory-MLP-Mixer (TMM) and Trajectory-Transformer (TT).

\begin{figure}[t] 
\centering 
\includegraphics[width=1\linewidth]{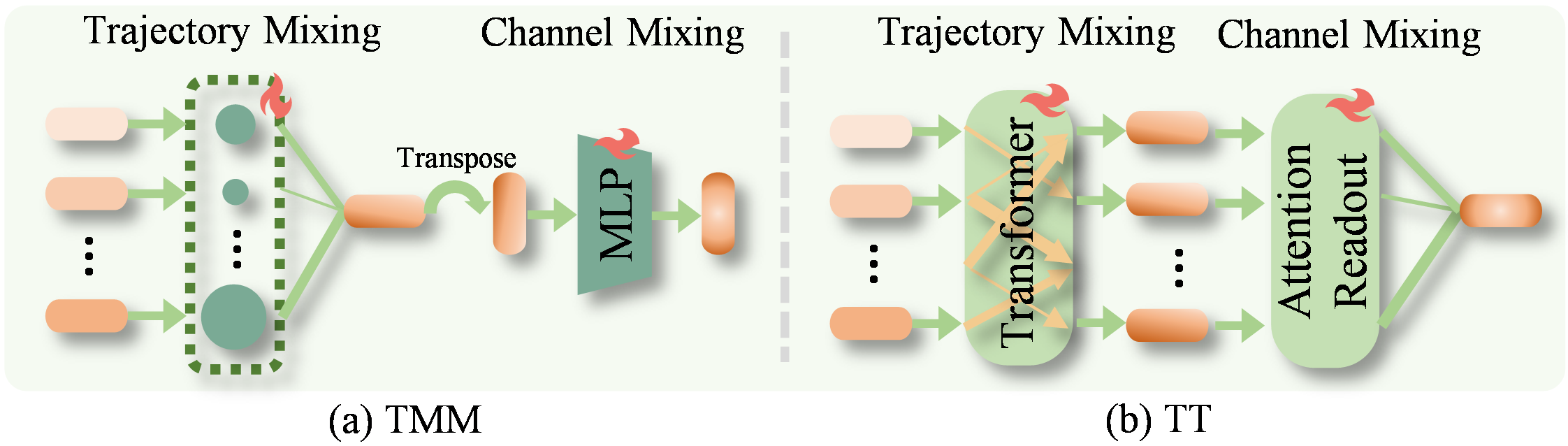} 
\caption{Instantiations of MVC Encoder. The left one is Trajectory-MLP-Mixer, the circle size denotes the weights of each hop information. The right denotes Trajectory-Transformer which treats each hop feature as input token. }
\label{mvc} 
\end{figure}

\textbf{1) Trajectory-MLP-Mixer (TMM)}. Considering simplicity and efficiency, we introduce a lightweight model, \ie, Trajectory-MLP-Mixer, to instantiate the MVC encoder as depicted in Figure~\ref{mvc}a. Initially, trajectory mixing is applied to the trajectory dimension to adaptively integrate information from each hop. Subsequently, channel mixing is applied to the mixed trajectories to obtain condensed multi-view features:
\begin{equation}
    \hat{T}=\mathcal{F}_\omega(T)=\operatorname{TMM}({T})=\sigma(w_{\text{traj}}\odot T)\oplus T W_{\text{channel}}
\end{equation}
where \( w_{\text{traj}} \in \mathbb{R}^{1 \times d} \) is employed for trajectory mixing, $\odot$ denotes the broadcasted Hadamard product with a summation across the feature dimension, and $\sigma$ represents the normalization function. Thus, $\sigma(w_{\text{traj}} \odot T) \in \mathbb{R}^{L \times N}$ signifies the importance weights for each hop information of each node, which will be utilized for the weighted summation $\oplus$ of multi-view information $T$. Additionally, \( W_{\text{channel}} \in \mathbb{R}^{d \times d} \) is used for channel mixing.

\textbf{2) Trajectory-Transformer (TT)}. 
We treat the trajectories as a sequence, with each hop information considered a token input, as illustrated in Figure~\ref{mvc}b. These tokens are fed into a Transformer encoder to learn the mixed-hop representations. Subsequently, an attention-based readout function is developed to adaptively aggregate mixed-hop features to obtain condensed trajectory features. The steps can be formulated as follows:
\begin{equation}
    \hat{T} =\mathcal{F}_\omega(T)= \operatorname{TT}(T)=\mathcal{R}(\operatorname{Tr}(T))
\end{equation}
where \(\operatorname{Tr}(\cdot)\) denotes a shallow Transformer encoder and \(\mathcal{R}\) represents an attention-based readout function.

\subsection{Adaptive Graph Upsampler}\label{sec:upsampler}
After obtaining condensed trajectory features $\hat{T}$, we use them to compute the upsampling probability for each existing edge to decelerate message passing, thus mitigating the over-smoothing problem (theoretical justification can be found in Section~\ref{sec:justification}). Subsequently, we employ a differentiable sampling method to generate an augmented graph. Specifically, we compute the upsampling logits as follows:
\begin{equation}
      P_{ij} = {W}_{\text{up}} \left[ \hat{{T}}_i \| \hat{{T}}_j \right],\, \forall (i,j) \in \mathcal{E},
\end{equation}
where \( P_{ij} \in \mathbb{R}^{2} \) represents the upsampling logits for the edge between nodes \( i \) and \( j \), \( {W}_{\text{up}} \) is a learnable weight matrix, and \( \left[ \hat{{T}}_i \| \hat{{T}}_j \right] \) denotes the concatenation of the condensed trajectory features of nodes \( i \) and \( j \).

Then we employ straight-through Gumbel-Softmax~\citep{gumbelsoftmax,bengio2013estimating} to derive the upsampling mask and generate the augmented graph $\hat{G}$:
\begin{gather}
    \hat{P}=\operatorname{softmax}({P+\epsilon_{\text{gumbel}}}) \in \mathbb{R}^{|\mathcal{E}|\times 2} \nonumber \\
    M=\mathds{1}_{\hat{P}[:,0]\leq \hat{P}[:,1]}-\operatorname{sg}(\hat{P})+\hat{P} \in \mathbb{R}^{|\mathcal{E}|}
\end{gather}
where \( \epsilon \in \mathbb{R}^{|\mathcal{E}| \times 2} \) represents Gumbel noise used to introduce diversity into the sampling process (exploration), preventing the selection of edges solely based on the highest probabilities and ensuring the differentiability. \( \hat{P} \) denotes the upsampling probability, and \( \mathds{1}_{\hat{P}[:,0] \leq \hat{P}[:,1]} \) is an indicator function that returns 1 if \( \hat{P}[:,0] \leq \hat{P}[:,1] \). \( \operatorname{sg(\cdot)} \) denotes the stop gradient operator. This approach achieves two objectives: (1) ensures that the output is exactly one-hot (by adding and then subtracting a soft value), and (2) maintains the gradient equal to the soft gradient (by keeping the original gradients).
\( M \) represents an active mask used to determine which edges will have a new node inserted. Specifically, if the mask for edge \( i \to j \) is 1, a new node \( k \) is inserted between nodes \( i \) and \( j \). Otherwise, the original edge \( i \to j \) is then masked and replaced by new edges \( i \to k \) and \( k \to j \). 
The initial feature vector of node \( k \) can be set to an adaptive sum: 
\begin{equation}
    \hat{X}_{k} = {P}_{ij}^{(l)}[0]{X}_{i} +{P}_{ij}^{(l)}[1]{X}_{j}
\end{equation}
Other feature initialization strategies, such as zero-initialization and weighted sum, can also be employed. Their empirical results are presented in Section~\ref{sec:ablation}.
Then, we obtain the augmented graph \(\hat{{G}}=(\hat{\mathcal{V}}, \hat{\mathcal{E}}, \hat{X})\), which will be used for downstream tasks to mitigate over-smoothing and improve overall performance.

\subsection{Model Training}\label{sec:training}
To jointly optimize the augmented graph and the downstream model \(\mathcal{D}_{\psi}(\cdot)\) through downstream tasks, such as node classification, the augmented graph will be input into the downstream model to compute the downstream loss $\mathcal{L}_{\text{down}}$. Additionally, we incorporate over-smoothing metrics, such as Mean Average Distance (MAD)~\citep{chen2019measuring} and Dirichlet Energy~\cite{rusch2023survey}, as regularization terms $\mathcal{L}_{\text{smooth}}$ to ensure that \our explicitly focuses on mitigating the over-smoothing issue. The total loss is defined as:
\begin{equation}
\begin{aligned}
    \mathcal{L}_{\text{total}}&=\mathcal{L}_{\text{down}}+\beta\mathcal{L}_{\text{smooth}}\\
    &=-\mathbb{E}_{v\in\hat{\mathcal{V}}}y\log p(\tilde{y}\mid v,\mathcal{N}_v)-\beta\operatorname{MAD}(\hat{G})
\end{aligned}
\end{equation}
where $\tilde{y}$ represents the prediction of the downstream model, $\mathcal{N}_v$ denotes the set of neighboring nodes associated with node $v$ used in the downstream model and $\beta$ is the regularization coefficient that controls the impact of the smoothing metric. More analysis about $\beta$ is detailed in the Appendix~\ref{appendixbeta}. In the experiments, we used the same settings for all baselines to ensure that the comparisons are fair. We empirically set $\beta$ to 1 because a larger value would negatively impact downstream performance. The formula of MAD is represented as:
\begin{equation}
\operatorname{MAD}(X) = \frac{1}{|\mathcal{V}|}\sum_{i \in \mathcal{V}}\sum_{j \in \mathcal{V}_{i}}1-\frac{X_{i}^{T}X_{j}}{||X_{i}||||X_{j}||} 
\end{equation}
Through optimizing the downstream loss, all parameters in our framework will be updated in an end-to-end manner through backpropagation. After warming up the downstream model, it will be used to compute trajectories instead of relying solely on the trajectory precomputation module mentioned in Section~\ref{sec:trace}. The model training process continues until convergence, aiming to obtain the optimal upsampled graph and model parameters.

\section{Theoretical Analysis and Discussion}\label{sec:discuss}
In this section, we will first explain \our as a universal framework in Section~\ref{sec:variants}. Then, we will provide a theoretical analysis of why \our works in Section~\ref{sec:justification}. Last, we will present the complexity analysis of \our in Section~\ref{sec:complexity}.

\subsection{UniGAP as A Universal Framework}\label{sec:variants}
\our is a universal plug-in that can be integrated with various GNN models, as demonstrated in Section~\ref{sec:overal_exp}. Additionally, UniGAP is a fully differentiable trajectory-based framework whose components, including the Trajectory Precomputation module, the MVC encoder, and the graph modification strategy, are all modular and can be replaced in different ways. In this subsection, we discuss its potential conceptual relationships with methods such as HalfHop and AdaEdge~\citep{azabou2023halfhop,adaedge}.

\textbf{HalfHop}~\cite{azabou2023halfhop}. 
With Multi-View Condensation and Graph Upsampler frozen with random parameters. The upsampling of nodes then becomes a stochastic process with probability \( p \) based on these parameters, making UniGAP function as a randomized heuristic upsampler, similar to HalfHop.

\textbf{AdaEdge}~\cite{adaedge}. 
With Trajectory Computation contain only the last layer's embedding of the downstream GNN, the upsampling probability for each edge in the Adaptive Graph Upsampler can be considered a binary classifier of intra-class or inter-class. This allows UniGAP adaptively increasing intra-class connections and decreasing inter-class connections, similar to AdaEdge.

\subsection{Why Does \our Work?}\label{sec:justification}
Equipped with \our, the optimized upsampled graph can mitigate the over-smoothing problem, thus enhancing downstream performance.
In this subsection, we will provide theoretical explanations of: (1) UniGAP can learn in concert with downstream tasks to get the optimal upsampling graph structure. (2) UniGAP is able to slow the over-smoothing problem.
These analyses are based on a simplified setting based on linear GNNs with mean aggregation, following prior theoretical work on over-smoothing. The goal of these analyses is to provide intuition about the smoothing dynamics induced by graph upsampling, rather than to exactly characterize the behavior of the non-linear architectures used in our experiments.

\noindent\textbf{Closed-Form Solution.}\quad
To simplify the analysis, we assume the GNN model as linear GNN with mean aggregation~\citep{theoreticoversmoothing}. Therefore, the $k$-th hop node features at $e$-th epoch are:
\begin{equation}
    \begin{aligned}
H^{(k)}_{e} &=\hat{A}^{k}_{e}X\\
        &=\operatorname{UniGAP}(H^{(k)}_{e-1},\cdots,H^{(0)}_{e-1},\theta_{u})^{k}X\\
        &:=f^{k}(A,\theta_{u},X)
    \end{aligned}
    \label{eq:node_feat}
\end{equation}
where $\hat{A}$ is the adaptive message passing matrix, and $\theta_{u}$ is the learnable parameters of UniGAP.
\begin{proposition}[]
By jointly training the parameters of the downstream model \(\theta\) and \our \(\theta_{u}\) at each epoch, \our learns the optimal structural representation for downstream tasks by optimizing \( \langle\hat{\theta}, \hat{\theta}_{u}\rangle\) as follows:
\begin{equation}
    \begin{aligned}
        &\langle\hat{\theta},\hat{\theta}_{u}\rangle_{e} = \argmin_{\langle\theta,\theta_{u}\rangle} \frac{1}{2n}||Y-H^{(k)}_{e}\theta||^{2}+\lambda||\theta||^{2} +\gamma||\theta_{u}||^{2}\\
        &\quad-\beta \operatorname{MAD}(H^{(k)}_{e})\\
&= \Bigg\langle\left( \frac{(f^{k})^{T}f^{k}}{n}+\lambda I \right)^{-1}\frac{(f^{k})^{T}Y}{n}, \\ &\quad\gamma^{-1} \left( \frac{1}{2n} \left( \frac{\partial f^{k}}{\partial \theta_u} \right)^T (Y - f^{k} \hat{\theta}) \hat{\theta} - \frac{\beta}{2} \frac{\partial}{\partial \theta_u} \text{MAD}(f^{k}) \right) \Bigg\rangle\\
    \end{aligned}
    \label{eq:pro}
\end{equation}
\end{proposition}
where the subscript $e$ means the $e$-th epoch. As shown in Equation~\ref{eq:node_feat}, \ref{eq:pro}, UniGAP can learn the message passing dynamics and adaptivly change the nodes' receptive filed, finally reach a optimal structural representation for downstream task. The derivation and more illustrations can be found in Appendix~\ref{appendixproof}.

\noindent\textbf{Slower Smoothing.}\quad We find that \our can decelerate the message passing process by slowing down the smooth rate of the covariance of node features during message passing. The following Lemma and Theorem shows the approximated covariance of the node features after $k$ rounds of message passing with or without \ours.

\begin{definition}
Following ~\citep{azabou2023halfhop, theoreticoversmoothing}, for a symmetric positive semi-definite matrix $S \in \mathbb{R}^{d \times d}$, we define the risk function
\begin{equation}
    \mathcal{R}_{\text{reg.}}(S) \overset{def_{.}}{=} (\Sigma^{\frac{1}{2}}\beta^{*})^{T}K(\Sigma^{\frac{1}{2}}\beta^{*}) \in \mathbb{R}_{+}
\end{equation}
where $K=(Id-S^{\frac{1}{2}}M(\gamma Id+M^{T}SM)^{-1}M^{T}S^{\frac{1}{2}})^{2}$, $\Sigma$ is the latent model covariance, $M$ is the projection matrix, $\beta^{*}$ are the true model parameters, and $\gamma$ is the ridge penalty in the least-squares estimator.
\end{definition}

\begin{lemma}\label{lemma3.3}
The risk after k rounds of Message Passing can be approximated as $\mathcal{R}^{(k)} \simeq \mathcal{R}_{\text{reg.}}(A^{2k}\Sigma)$, where $A = (I+\Sigma^{-1})^{-1}$. 
\end{lemma}
As shown in Lemma~\ref{lemma3.3}, the rate of smoothing of original MP is $2k$ due to the $A^{2k}\Sigma$ operator~\citep{theoreticoversmoothing}. Then we consider the condition with \ours.
\begin{lemma}\label{lemma3.4}
With UniGAP, the risk of test set $\mathcal{R}_{\text{reg.}}(\Sigma^{(k)}_{\text{UniGAP}})$ after $k$ rounds of message passing with UniGAP can be approximated as:
\begin{equation}
    \mathcal{R}_{\text{reg.}}(\Sigma^{(k)}_{\text{UniGAP}}) \simeq \mathcal{R}_{\text{reg.}}(\frac{1}{2}A^{k-1}(I+((1-p)I+pA)^{2})\Sigma)
\end{equation}
\end{lemma}
Here $\Sigma^{(k)}_{\text{UniGAP}}$ is the approximated covariance of the node features, $p = \mathbb{E}(\hat{P}[:,0])$ and $\hat{P}$ denotes the upsampling probability at the optimal graph structure. The rate of smoothing of MP with UniGAP is $k-1$ due to the $A^{k-1}\Sigma$ operator. The complete proof can be found in Appendix~\ref{appendixproof}.

With Lemma~\ref{lemma3.3} and \ref{lemma3.4}, we can naturally get the Theorem:
\begin{theorem}[Slower Smoothing of \our]\label{theo:smooth}
    After $k$ rounds of message passing, the risk obtained with UniGAP is  $\mathcal{R}^{(k)}_{\text{UniGAP}} \propto A^{k-1}$, roughly half the rate of smoothing compared to original MP model's $\mathcal{R}^{(k)} \propto A^{2k}$.
\end{theorem}
By slowing down the smoothing rate of the covariance of node features during message passing from $2k$ to $k-1$, UniGAP roughly half the over-smoothing problem.

\subsection{Complexity Analysis}
\label{sec:complexity}
In this subsection, we will present a complexity analysis of \our from a theoretical perspective. 

\textbf{Time Complexity.}
Our framework comprises four primary components: trajectory computation, multi-view condensation, adaptive graph upsampler, and downstream model. The trajectory precomputation module involves complexities of $\mathcal{O}(1)$, $\mathcal{O}(LEd)$, or $\mathcal{O}(LEd+LNd^2)$ depending on the initialization method: zero initialization, non-parametric message passing, or pre-trained GNN. The MVC Condensation module (Trajectory-MLP-Mixer) operates with a time complexity of $\mathcal{O}(LN+Nd^2)$, while the adaptive graph upsampler computes with a complexity of $O(Ed)$ to determine upsampling probabilities for adjacent node pairs. 
The time complexity of downstream model, \eg, GNN, is $\mathcal{O}(LEd+LNd^2)$.
Consequently, the total time complexity of our method is $\mathcal{O}(LN+LEd+LNd^2)$, comparable to traditional L-layer GNNs, \ie, $\mathcal{O}(LEd+LNd^2)$.

\textbf{Space Complexity.}
The space complexity of our method also closely aligns with that of traditional GNNs ($\mathcal{O}(E + Ld^2 + LNd)$). Specifically, the trajectory precomputation module exhibits space complexities of $\mathcal{O}(1)$, $\mathcal{O}(E)$, or $\mathcal{O}(E + Ld^2 + LNd)$. The multi-view condensation module requires $\mathcal{O}(d + d^2 + LN + Nd)$, whereas the adaptive graph upsampler operates with a complexity of $\mathcal{O}(d + E)$. The space complexity of the downstream model, such as GNN, is $\mathcal{O}(E + Ld^2 + LNd)$. Consequently, the total space complexity of our method is $\mathcal{O}(d + d^2 + LN + Nd + E + Ld^2 + LNd)$ which is similar to that of GNN.

In summary, the complexity of \our is on par with traditional GNN methods, maintaining efficiency.

\section{Experiments}
In this section, we present a comprehensive empirical investigation. Specifically, we aim to address the following research questions: 
\textbf{RQ1}: How does the proposed \our approach perform when applied to various GNN models across both homophilic and heterophilic datasets?
\textbf{RQ2}: To what extent can UniGAP effectively mitigate the phenomenon of over-smoothing? 
\textbf{RQ3}: How does \our mitigate over-smoothing? 
\textbf{RQ4}: How do the various strategic choices across different modules impact performance?
\textbf{RQ5}: Can we further enhance \our with advanced techniques like LLM?

\subsection{Experimental Setup}
\textbf{Datasets.}\quad 
We evaluated our method on 11 node classification benchmark datasets, as shown in Table~\ref{table1} and Table~\ref{tablelargescale}, comprising 4 homophilic (1 large-scale) and 7 heterophilic datasets (4 large-scale). More detailed information is provided in Appendix~\ref{appendixexp}.

\textbf{Evaluation Protocols.}\quad
To compare \our with baselines, we use public split setting for all datasets, and report the mean accuracy along with the standard deviation across twenty different random seeds. \our is integrated as a plug-in method with various SOTA GNN models to evaluate its effectiveness. 
Besides, we report the MAD metric to observe the phenomenon of over-smoothing as the number of layers increases.

\textbf{Baselines.}\quad
The baselines primarily consist of four categories: (1) traditional GNN methods such as \textbf{GCN}~\citep{kipf2017semisupervised}, \textbf{GAT}~\citep{gat}, and \textbf{GraphSAGE}~\citep{graphsage}; (2) advanced GNN methods including \textbf{GCNII}~\citep{gcnii}, \textbf{GGCN}~\citep{ggcn}, \textbf{H2GCN}~\citep{h2gcn}, \textbf{GPRGNN}~\citep{gprgnn}, \textbf{AERO-GNN}~\citep{aerognn}, DRGCN~\citep{dr} and MTGCN~\citep{mtgcn}; (3) Graph Transformer methods like \textbf{GT}~\citep{gt}, \textbf{GraphGPS}~\citep{rampášek2023recipe}, and \textbf{NAGphormer}~\citep{chen2023nagphormer}; and (4) other graph upsampling methods, \ie, \textbf{AdaEdge}, \textbf{HalfHop} and non-upsampling methods such as \textbf{DropEdge}~\citep{rong2020dropedge}. We integrate \our, DropEdge, HalfHop and AdaEdge with various GNN models for evaluation. 

\subsection{Overall Performance Comparison (RQ1)}\label{sec:overal_exp}
\subsubsection{Experiments on Small and Medium-scale Datasets}
For heterophilic datasets, \our significantly enhances traditional GNN models, which heavily rely on the ``homophily'' assumption. For example, \our achieves an average absolute improvement on four heterophilic datasets of 10.0\% on GCN, 13.3\% on GAT, and 5.0\% on GraphSAGE, respectively. It also enhances other advanced GNN models, with absolute improvements such as 2.3\% on GCNII, 1.1\% on H2GCN, and 2.9\% on GPRGNN.
Compared to HalfHop, \our surpasses it in all heterophilic datasets with over 3.9\% average absolute improvement with GCN, demonstrating the effectiveness of adaptive graph upsampling in identifying the optimal graph structure to enhance the downstream model's performance. 

For homophilic datasets, traditional GNN models benefit from their inductive bias, specifically the homophily assumption, leading to moderate performance. However, UniGAP can still identify graph structures that enhance downstream tasks through adaptive learning, thereby improving the performance of all baselines across different datasets. Average absolute improvements range from 1.1\% to 4.8\%.

In summary, compared to all the baselines and datasets, UniGAP demonstrates a significant improvement, with an average absolute boost ranging from 1.4\% to 6.7\%, surpassing the SOTA baselines, showcasing our effectiveness.

\begin{table*}[t]
\caption{Node classification performance on homophilic and heterophilic graphs. The results equipped with UniGAP are highlighted by a gray background. The boldface and underscore show the best and the runner-up,  respectively.}
\label{table1}
\begin{center}    
\renewcommand\arraystretch{0.943}
\resizebox{1.0\linewidth}{!}{
\begin{tabular}{lcccccccccc}
\toprule
Type      & \multicolumn{4}{c}{Homophilic graphs}                  & & \multicolumn{4}{c}{Heterophilic graphs}            \\ \cline{2-5} \cline{7-10}
Dataset       & Pubmed       & Citeseer     & Cora         &Arxiv&         & Texas       & Cornell      & Wisconsin   & Actor      &    \\
Homophily     & 0.66         & 0.63         & 0.77         &0.42 &         & 0.00        & 0.02         & 0.05        & 0.01       &Avg \\
\midrule

GT~\citep{gt}                            & 79.08 ± 0.4  & 70.16 ± 0.8  & 82.22 ± 0.6  & 70.63 ± 0.4  && 84.18 ± 5.4  & 80.16 ± 5.1  & 82.74 ± 6.0 & 34.28 ± 0.7  &72.92 \\
GraphGPS~\citep{rampášek2023recipe}      & 79.94 ± 0.3  & 72.43 ± 0.6  & 82.44 ± 0.6  & 70.97 ± 0.4  && 82.21 ± 6.9  & 82.06 ± 5.1  & 85.36 ± 4.2 & 36.01 ± 0.9  &73.92 \\
NAGphormer~\citep{chen2023nagphormer}    & 80.57 ± 0.3  & 72.75 ± 0.8  & 84.20 ± 0.5  & 70.13 ± 0.6  && 80.12 ± 5.5  & 79.89 ± 7.1  & 82.97 ± 3.2 & 34.24 ± 0.9  &73.11 \\
AERO-GNN~\citep{aerognn}                 & 80.59 ± 0.5  & 73.20 ± 0.6  & 83.90 ± 0.5  & 72.41 ± 0.4  && 84.35 ± 5.2  & 81.24 ± 6.8  & 84.80 ± 3.3 & 36.57 ± 1.1  &74.63 \\
DRGCN~\citep{dr}        & 80.16 ± 0.5  & 73.35 ± 0.4  &84.12 ± 0.6   &73.27± 0.3   &&85.07 ± 4.4   & 84.69 ± 5.1  &\underline{87.71  ± 4.5}  &36.03 ± 1.2   &75.55 \\
MTGCN~\citep{mtgcn}     & \underline{80.61 ± 0.4}   & 72.74 ± 0.3 & 83.35 ± 0.4   &73.52 ± 0.4   && 85.16 ± 6.1 &84.37 ± 4.9  &86.04 ± 3.5  &36.83 ± 1.1   &75.32 \\
\midrule
GCN~\citep{kipf2017semisupervised}           & 79.54 ± 0.4  & 72.10 ± 0.5  & 82.15 ± 0.5  & 71.74 ± 0.3  && 65.65 ± 4.8 & 58.41 ± 3.3  & 62.02 ± 5.9 & 30.57 ± 0.7   &65.27 \\
\  +DropEdge~\citep{rong2020dropedge}        & 78.16 ± 0.6  & 72.12 ± 0.6  & 82.21 ± 0.5  & 71.87 ± 0.2  && 67.01 ± 4.0 & 58.74 ± 4.5  & 66.59 ± 6.1 & 30.72 ± 0.7   &65.93 \\
\  +AdaEdge~\citep{adaedge}                  & 78.64 ± 0.7  & 72.15 ± 0.4  & 81.74 ± 0.4  & 71.98 ± 0.3  && 67.34 ± 4.2 & 62.08 ± 3.6  & 66.79 ± 6.5 & 30.87 ± 0.6   &66.45 \\
\  +HalfHop~\citep{azabou2023halfhop}        & 77.42 ± 0.6  & 72.40 ± 0.7  & 81.36 ± 0.6  & 72.59 ± 0.2  && 71.62 ± 3.9 & 72.39 ± 4.8  & 66.14 ± 6.6 & 30.69 ± 0.7   &68.07 \\
\rowcolor{lightgray}
\  +\our(Ours)                                     & 80.06 ± 0.6  & 73.15 ± 0.6  & 84.20 ± 0.5  & 72.97 ± 0.2  && 77.37 ± 3.5 & 75.95 ± 4.7  & 69.59 ± 6.2 & 33.73 ± 0.9   &70.87 \\
\midrule
GAT~\citep{gat}                              & 77.81 ± 0.4  & 70.89 ± 0.8  & 83.18 ± 0.5  & 71.95 ± 0.4  && 60.46 ± 6.2  & 58.22 ± 3.7  & 63.59 ± 6.1 &30.36 ± 0.9   &64.57 \\
\  +DropEdge                                 & 77.15 ± 0.7  & 71.04 ± 0.9  & 83.20 ± 0.7  & 71.92 ± 0.5  && 69.45 ± 5.1  & 62.41 ± 5.6  & 68.47 ± 6.6 &31.05 ± 0.8   &66.84 \\
\  +AdaEdge                                  & 75.24 ± 0.6  & 70.92 ± 0.9  & 82.67 ± 0.7  & 72.08 ± 0.2  && 75.04 ± 7.4  & 69.23 ± 5.3  & 71.52 ± 6.7 &30.49 ± 0.9  &68.40 \\
\  +HalfHop                                  & 78.03 ± 0.5  & 71.83 ± 0.8  & 82.29 ± 0.8  & 72.56 ± 0.4  && 74.54 ± 6.1  & 66.48 ± 5.5  & 70.17 ± 6.0 &30.28 ± 1.0   &68.27 \\
\rowcolor{lightgray}
\  +\our(Ours)                  & 80.10 ± 0.3  & 73.20 ± 0.5  & 84.23 ± 0.6  & \textbf{74.06 ± 0.3}  && 78.64 ± 4.9  & 76.04 ± 5.1  & 74.81 ± 5.3 &32.79 ± 0.8   &71.73 \\
\midrule
GraphSAGE~\citep{graphsage}     & 78.67 ± 0.4  & 71.85 ± 0.6  & 83.76 ± 0.5  & 71.49 ± 0.3  && 82.43 ± 6.1  & 75.95 ± 5.3  & 81.18 ± 5.5 & 34.23 ± 1.0  &72.44 \\
\  +DropEdge                    & 78.89 ± 0.7  & 71.72 ± 0.5  & 83.58 ± 0.6  & 71.42 ± 0.5  && 83.94 ± 5.7  & 77.42 ± 6.1  & 82.44 ± 4.9 & 34.75 ± 0.9  &73.02 \\
\  +AdaEdge                     & 79.02 ± 0.7  & 71.07 ± 0.7  & 83.69 ± 0.6  & 70.24 ± 0.6  && 83.42 ± 5.9  & 79.81 ± 7.2  & 84.69 ± 5.4 & 34.97 ± 0.8  &73.36 \\
\  +HalfHop                     & 78.94 ± 0.3  & 72.07 ± 0.8  & 84.27 ± 0.6  & 72.93 ± 0.5  && 85.95 ± 6.4  & 74.60 ± 6.0  & 85.88 ± 4.0 & 36.82 ± 0.7  &73.92 \\
\rowcolor{lightgray}
\  +\our(Ours)                        & 80.21 ± 0.7  & \underline{73.60 ± 0.3}  & \textbf{86.46 ± 0.4}  & \underline{73.71 ± 0.2}  && \textbf{86.52 ± 4.8}  & 83.21 ± 5.7  & 86.98 ± 3.2 & 37.14 ± 0.8  &\textbf{75.98} \\
\midrule
GCNII~\citep{gcnii}         & 80.14 ± 0.7  & 72.80 ± 0.5  & 84.33 ± 0.5  & 72.74 ± 0.2  && 78.59 ± 6.6  & 78.84 ± 6.6  & 81.41 ± 4.7 & 35.76 ± 1.0  &73.08 \\
\  +DropEdge                & 80.18 ± 0.9  & 72.49 ± 0.7  & 84.74 ± 0.5  & 72.80 ± 0.7  && 79.55 ± 6.2  & 79.23 ± 6.5  & 80.01 ± 4.7 & 35.85 ± 1.0  &73.11 \\
\  +AdaEdge                 & 79.62 ± 0.7  & 72.60 ± 0.6  & 85.07 ± 0.7  & 70.62 ± 0.4  && 78.09 ± 7.2 & 76.54 ± 4.8  & 80.47 ± 4.0  & 36.11 ± 0.9  &72.39 \\
\  +HalfHop                 & 78.96 ± 0.9  & 72.43 ± 0.7  & 84.91 ± 0.6  & 72.91 ± 0.3  && 79.84 ± 6.1  & 80.04 ± 5.1  & 82.96 ± 3.7 & 36.01 ± 0.9  &73.50 \\
\rowcolor{lightgray}
\  +\our(Ours)                    & \textbf{80.79 ± 0.7}  & \textbf{73.70 ± 0.4}  & \underline{86.22 ± 0.4}  & 73.32 ± 0.3  && 82.08 ± 4.2  & 81.16 ± 5.0  & 84.06 ± 3.8 & 36.47 ± 1.1  &74.76 \\
\midrule
GGCN~\citep{ggcn}          & 78.34 ± 0.6  & 71.60 ± 0.6  & 83.64 ± 0.6  & 71.29 ± 0.1  && 82.86 ± 4.5  & \textbf{85.68 ± 6.6}  & 86.86 ± 3.2 & \underline{37.54 ± 1.5}  &74.73 \\
\  +DropEdge               & 77.28 ± 0.9  & 70.80 ± 0.5  & 83.84 ± 0.8  & 71.32 ± 0.4  && 81.57 ± 4.6  & 82.72 ± 7.1  & 87.04 ± 3.6 & 35.11 ± 0.9  &73.71 \\
\  +AdaEdge                & 76.54 ± 1.2  & 70.33 ± 0.9  & 79.86 ± 0.9  & 69.24 ± 0.5  && 80.66 ± 4.3 & 81.37 ± 7.9  & 87.24 ± 3.0 & 35.52 ± 1.1   &72.60 \\
\  +HalfHop                & 77.92 ± 0.8  & 70.90 ± 0.8  & 84.07 ± 0.7  & 71.12 ± 0.3  && 83.46 ± 5.1  & 83.22 ± 7.1  & 86.49 ± 4.6 & 36.32 ± 1.3  &74.19 \\
\rowcolor{lightgray}
\  +\our(Ours)                   & 80.09 ± 0.7  & 73.15 ± 0.6  & 84.97 ± 0.6  & 73.08 ± 0.3  && 85.14 ± 4.9  & 84.27 ± 6.0  & 87.69 ± 3.3 & \textbf{37.69 ± 1.2}  &75.74 \\
\midrule
H2GCN~\citep{h2gcn}         & 78.12 ± 0.4  & 71.80 ± 0.8  & 84.26 ± 0.6  & 70.97 ± 0.2  && 84.86 ± 7.2  & 82.70 ± 5.2  & 87.65 ± 4.9 & 35.70 ± 1.0  &74.51 \\
\  +DropEdge                & 78.32 ± 0.8  &71.95 ± 0.9   & 83.17 ± 0.7  & 71.02 ± 0.6  && 85.11 ± 7.1  & 82.92 ± 5.1  & 87.91 ± 4.7 & 35.71 ± 1.2  &74.51 \\
\  +AdaEdge                 & 78.29 ± 0.7  & 71.65 ± 0.7  & 83.31 ± 0.6  & 70.28 ± 0.5  && 83.79 ± 7.4 & 83.02 ± 5.0  & 87.38 ± 4.7 & 35.77 ± 1.1   &74.19 \\
\  +HalfHop                 & 79.23 ± 0.8  & 71.89 ± 0.9  & 82.65 ± 0.8  & 71.26 ± 0.3  && 85.02 ± 7.0  & 83.54 ± 5.9  & 86.22 ± 5.2 & 34.83 ± 1.2  &74.33 \\
\rowcolor{lightgray}
\  +\our(Ours)                    & 80.26 ± 0.6  & 73.10 ± 0.5  & 85.13 ± 0.6  & 73.19 ± 0.2  && \underline{86.14 ± 6.5}  & \underline{84.96 ± 5.0}  & \textbf{87.73 ± 4.8} & 36.42 ± 0.8  &\underline{75.87} \\
\midrule
GPRGNN~\citep{gprgnn}        & 75.68 ± 0.4  & 71.60 ± 0.8  & 84.20 ± 0.5  & 71.86 ± 0.3  && 81.51 ± 6.1  & 80.27 ± 8.1  & 84.06 ± 5.2 & 35.58 ± 0.9  &73.10 \\
\  +DropEdge                 & 76.85 ± 1.1  & 71.74 ± 0.8  & 83.07 ± 0.7  & 71.09 ± 0.4  && 82.13 ± 6.7  & 82.92 ± 7.5  & 84.88 ± 6.0 & 35.21 ± 1.0  &73.49 \\
\  +AdaEdge                  & 78.03 ± 1.1  & 70.82 ± 0.7  & 81.76 ± 0.8  & 70.99 ± 0.3  && 81.97 ± 6.2 &  84.32 ± 7.2  & 83.64 ± 6.6 & 35.06 ± 0.7  &73.32 \\
\  +HalfHop                  & 77.95 ± 1.0  & 72.37 ± 0.7  & 83.49 ± 0.8  & 71.44 ± 0.5  && 83.87 ± 6.0  & 84.79 ± 7.6  & 85.76 ± 6.1 & 35.46 ± 1.0  &74.38 \\
\rowcolor{lightgray}
\  +\our(Ours)                     & 80.21 ± 0.9  & 72.90 ± 0.6  & 85.18 ± 0.4  & 71.98 ± 0.4  && 85.65 ± 5.4  & 84.37 ± 6.3  & 86.57 ± 5.0 & 36.51 ± 0.9  &75.42 \\

\bottomrule
\end{tabular}
}
\end{center}
\end{table*}

\subsubsection{Experiments on Large-scale Heterophilic Datasets}
To evaluate the performance of \our on large heterophilic datasets, we conducted experiments on Squirrel, Amazon-ratings, and Questions. As shown in Table \ref{tablelargescale}, comparing with 2 scalable GNNs and NAGphormer, \our achieves the best performance on all datasets, surpasses the runner-up over 1\% absolute improvement on Amazon2M, demonstrating its effectiveness for large-scale graphs.

This further validates the effectiveness of decoupling multi-view attention and message interaction in capturing information and learning implicit long-range dependencies at a large scale among numerous nodes. 

\begin{table}[t]
\caption{UniGAP maintains its superiority on large heterophilic datasets, demonstrating strong generalization capabilities.}
\label{tablelargescale}
\begin{center}    
\renewcommand\arraystretch{1}
\resizebox{1.0\linewidth}{!}{
\begin{tabular}{lccccc}
\toprule
Dataset     & Squirrel        & Amazon-ratings & Questions \\
Homophily   & 0.03            & 0.14           & 0.02      \\
Metric      & Accuracy        & Accuracy       & ROC-AUC   \\

\midrule
GCN                   &45.87±2.3  &53.80±0.6  &79.02±0.6 \\
\quad + AdaEdge       &54.79±1.8  &56.92±1.1  &80.85±0.8 \\
\quad + Halfhop       &60.24±2.4  &55.19±1.6  &79.60±1.1 \\
\rowcolor{lightgray}
\quad + UniGAP(Ours)  &65.58±1.6  &59.76±0.8  &82.73±0.9 \\
\midrule
GraphSAGE             &43.78±1.9  &55.40±0.2  &77.21±1.3 \\
\quad + AdaEdge       &55.29±1.2  &58.84±1.3  &80.17±1.0 \\
\quad + Halfhop       &54.83±2.7  &58.97±1.5  &78.62±1.7 \\
\rowcolor{lightgray}
\quad + UniGAP(Ours)  &65.32±1.4  &62.83±0.9  &84.21±0.8 \\
\midrule
H2GCN                 &43.62±2.3  &56.81±0.9  &79.28±0.8 \\
\quad + AdaEdge       &54.79±1.8  &58.77±1.2  &81.34±0.7 \\
\quad + Halfhop       &60.24±2.4  &59.61±1.0  &78.96±1.9 \\
\rowcolor{lightgray}
\quad + UniGAP(Ours)  &69.67±1.6  &63.75±1.1  &83.59±0.7 \\
\bottomrule
\end{tabular}
}
\end{center}
\end{table}

\subsection{The Degree of Over-smoothing (RQ2)}
To analyze the extent to which UniGAP effectively mitigates the phenomenon of over-smoothing, we conducted experiments on GCN across several datasets, evaluating its accuracy and MAD as the number of layers increases. As depicted in Figure~\ref{mad_cora}, the experimental results demonstrate that UniGAP significantly enhances model performance and alleviates over-smoothing at each layer, surpassing AdaEdge and HalfHop.

Specifically, while HalfHop and AdaEdge can mitigate over-smoothing to some extent, they do not achieve optimal performance as UniGAP does. HalfHop fails to identify crucial elements necessary to mitigate over-smoothing and enhance performance because it inserts nodes randomly rather than adaptively finding optimal positions for the inserted nodes. AdaEdge increases connectivity between intra-class nodes, but effective message propagation between inter-class nodes requires an optimal graph structure. Therefore, adaptive upsampling represents a more sophisticated approach to mitigating over-smoothing and enhancing model performance.

Moreover, we observe that adding \our on the baseline GCN substantially slows down the decay rate of MAD, roughly to about half of the original rate. For example, with \our, the MAD value of at the 6 layer is comparable to the smoothness level of the baseline GCN at the 3 layer, which is consistent with the theoretical analysis in Section~4.2. On heterophilic datasets such as Texas and Cornell, the effect of \our is more pronounced than on the homophilic Cora dataset, because heterophilic graphs are more susceptible to over-smoothing and therefore benefit more from slower propagation and a better adapted graph structure.

\subsection{Interpretability in Mitigating Over-smoothing (RQ3)}
To better analyze the reasons for mitigating over-smoothing, we examine the ratio of intermediate node insertions on a wide range of datasets. This analysis provides valuable insights into the root causes of over-smoothing. As shown in Figure~\ref{change}, when \our learned the augmented graph that achieved optimal performance, we identify the locations where intermediate nodes are added. We observe that new nodes tend to be inserted between inter-class node pairs, especially on heterophilic graphs where inter-class edges are more prevalent. This behavior slows down message passing across class boundaries and helps prevent the features of different classes from collapsing into similar vectors, thereby mitigating over-smoothing.

We also observe that a non-negligible fraction of new nodes is inserted on intra-class edges. One hypothesis is that some of these intra-class edges may in fact be noisy or weakly informative connections, and that inserting intermediate nodes on such edges effectively attenuates their influence and can lead to performance gains. A more systematic investigation of this hypothesis is left as a direction for future work. Overall, \our highlights key structural bottlenecks where over-smoothing tends to occur, offering an interpretable view of how adaptive upsampling reshapes the graph to alleviate this phenomenon.

\begin{figure}[t] 
\centering 
\includegraphics[width=1\linewidth]{fig/fig3_mad.png} 
\caption{The accuracy and MAD values of various upsampling methods in different layers on the Cora dataset.}
\label{mad_cora} 
\end{figure}

\begin{figure}[t] 
\centering 
\includegraphics[width=1\linewidth]{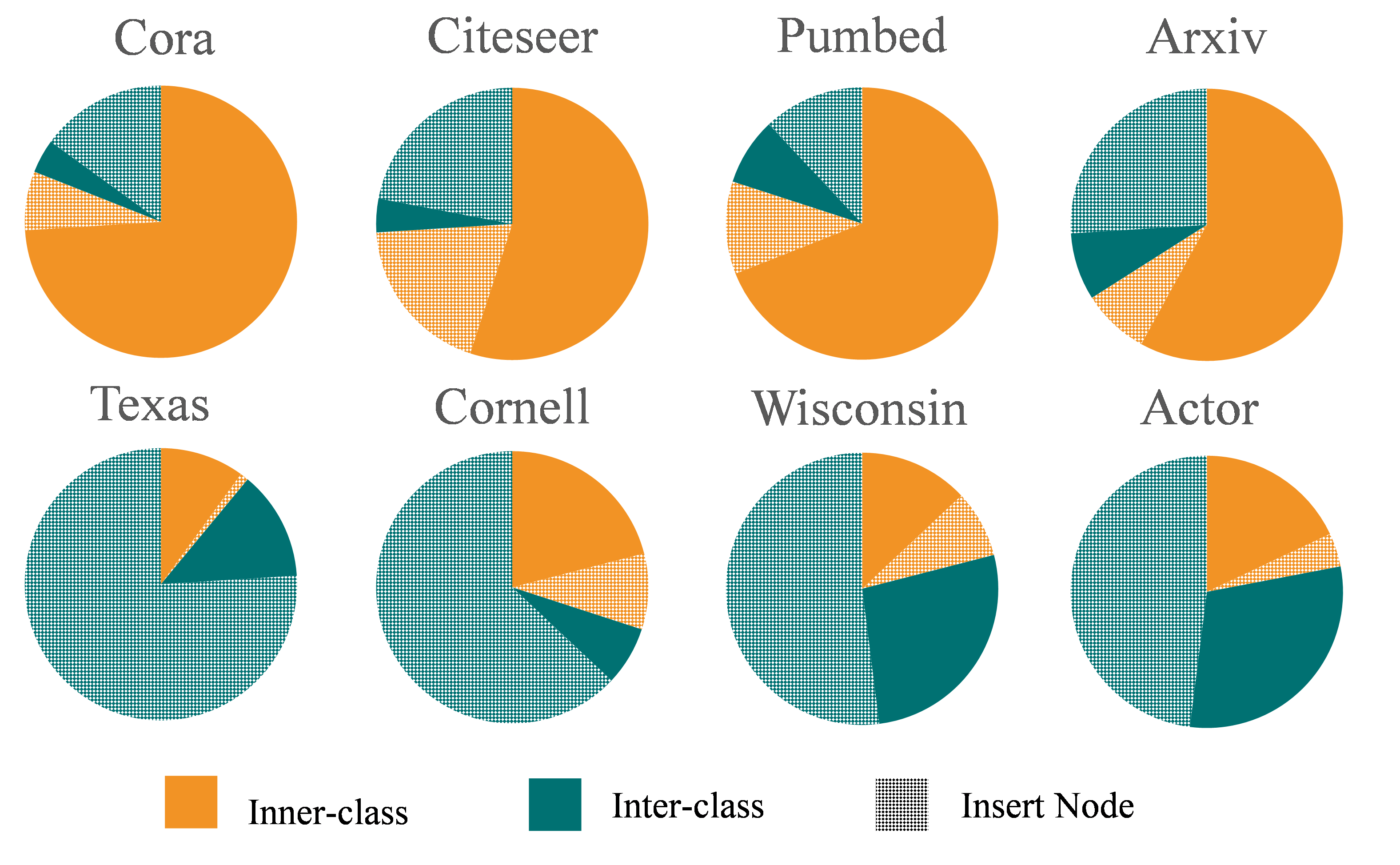} 
\caption{The proportion of inserted nodes for intra-class and inter-class edges by UniGAP of the optimal graph.}
\label{change} 
\end{figure}

\subsection{Ablation Study (RQ4)}\label{sec:ablation}
In Section~\ref{sec:trace}, \ref{sec:mvc} and \ref{sec:upsampler}, we provide various strategies for instantiating our modules. In this subsection, we will explore the impact of different strategies within the various modules of UniGAP on model performance.

\begin{figure}[t] 
\centering 
\includegraphics[width=1\linewidth]{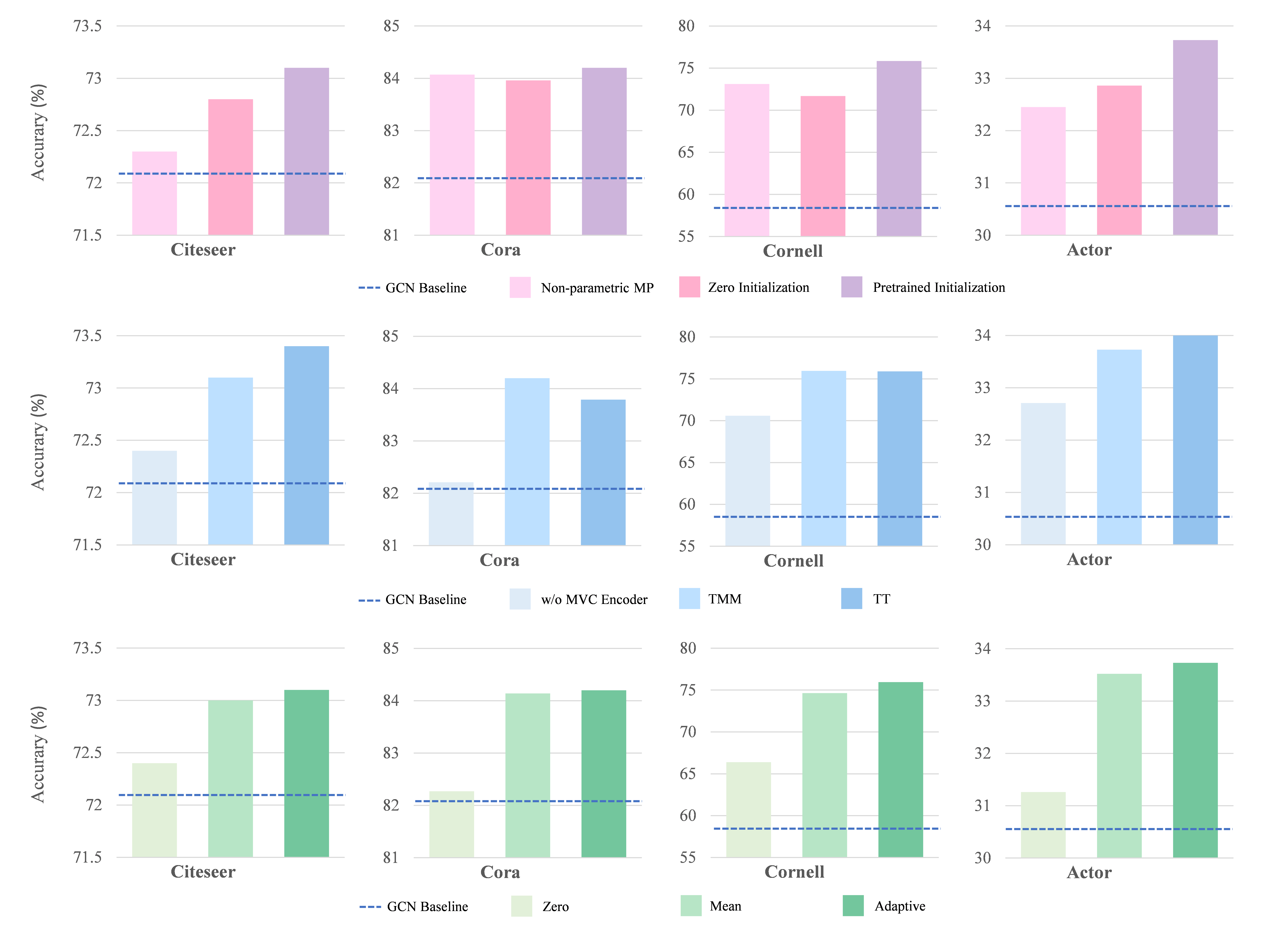} 
\caption{The performance of UniGAP with different Trajectory precomputation strategies (pink color), with different MVC Encoder (blue color), and with different initialization for inserted nodes (green color).}
\label{ab1} 
\end{figure}

\noindent\textbf{Comparison of Different Strategies in Trajectory Precomputation.}
In this study, we investigate three readily available strategies in Section~\ref{sec:trace}: Zero Initialization, Non-parametric Message Passing, and Pretrained GNN. We conduct experiments on GCN+\our across several datasets. From Figure~\ref{ab1}, we observe that Pretrained GNN consistently outperforms both Zero Initialization and Non-parametric Message Passing. Due to its larger number of parameters, it can learn unique, additional information from the graph, allowing UniGAP to optimize the graph structure from a more advantageous starting point.

Moreover, we conduct a sensitivity analysis on the pretrained GNN, where we adopt a weaker pretraining setup. We quantify the quality of the pretrained encoder by its validation performance on the pretext task. As shown in Table~\ref{pretraingnn}, better pretraining consistently improves the downstream accuracy of UniGAP. However, even with very weak or no pretraining, UniGAP still outperforms purely heuristic baselines, suggesting that UniGAP is not overly fragile to the quality of the initial trajectories.

\begin{table}[t]
\centering
\caption{Sensitivity of UniGAP to pretraining quality on the Texas dataset. The pretrain score is the validation performance on the pretext task (higher is better).}
\label{pretraingnn}
\renewcommand\arraystretch{1}
\resizebox{1.0\linewidth}{!}{
\begin{tabular}{lcc}
\toprule
Pretraining setup        & Pretrain score & UniGAP+GCN Accuracy \\
\midrule

None (Zero init)         & --    & 73.12±3.8 \\
Weak pretraining (10 ep.)  & 0.61  & 75.28±2.9 \\
Moderate pretraining (50 ep.)  & 0.72  & 75.66±2.7 \\
Strong pretraining (100 ep.) & 0.78  & 75.73±2.7 \\
Best pretraining (130 ep.) & 0.81  & 75.85±2.1 \\
Heuristic baseline (AdaEdge) & --    & 62.08±3.6 \\
\bottomrule
\end{tabular}
}
\end{table}

\noindent\textbf{Comparsion of Different Strategies in MVC Encoder.}
We investigate the MVC encoder using three strategies in this work: No encoder, Trajectory-MLP-Mixer (TMM), and Trajectory-Transformer (TT). As shown in Figure 4, both TMM and TT demonstrate significant improvements compared to not using an MVC Encoder, with absolute improvements ranging from 1\% to 1.92\%. Additionally, the MVC Encoder effectively compresses the trajectory features, reducing the time consumption for subsequent processes. Between TMM and TT, different datasets prefer different structures. For small datasets like Cora, we recommend using TMM as it is more lightweight than TT, thus avoiding the overfitting problem.

Moreover, to verify whether the MVC encoder is truly learning the propensity to over-smoothing from trajectories rather than simply extracting generic features that are useful for the downstream classification task, we design a diagnostic experiment. Specifically, we replace the trajectory inputs with only the initial node features and keep the rest of the MVC and UniGAP pipeline unchanged. As shown in Table~\ref{mvcab}, this modification leads to a clear performance degradation compared to using full trajectories, indicating that the MVC encoder is indeed leveraging information about the evolution of node representations across layers, rather than acting as a generic feature extractor.

\begin{table}[t]
\centering
\caption{Diagnostic experiment on the role of trajectories in the MVC encoder. We compare using full layer-wise trajectories versus using only the initial node features as input to MVC, while keeping the rest of the GCN+UniGAP pipeline unchanged.}
\label{mvcab}
\renewcommand\arraystretch{1}
\begin{tabular}{lccc}
\toprule
Dataset  & MVC Input & Test Accuracy \\
\midrule
Cora   & Trajectories        & 84.20±0.5 \\
       & Initial features only      & 81.37±0.7 \\
\midrule
Texas  & Trajectories        & 77.37±3.5 \\
       & Initial features only      & 71.52±3.7 \\
\bottomrule
\end{tabular}
\end{table}

\noindent\textbf{Comparsion of Different Initialization for Inserted Nodes.}
In Section~\ref{sec:upsampler}, we introduce three strategies for initializing features for inserted nodes: (1) zero initialization, (2) mean feature of source and target node, and (3) learnable weighted sum of source and target node features. As shown in Figure 4, the learnable weighted sum achieves the best performance. 
We hypothesize that this is because it encodes a natural inductive bias for node insertion: 
the feature of an intermediate node should lie between its two endpoints in the representation space, 
but the optimal position is not necessarily the arithmetic mean. 
By learning asymmetric weights for the two endpoints, the model can adaptively shift the inserted node closer to the more informative side while still preserving local semantic consistency and requiring only a small number of additional parameters. 
In contrast, zero initialization places the new node far from the data manifold, making it harder for the downstream GNN to integrate it meaningfully, and simple averaging ignores potential asymmetry between the connected nodes.

Beyond these simple choices, more expressive initialization strategies are also possible. 
For example, one could treat the features of inserted nodes as independent learnable embeddings. We leave a systematic exploration of these richer generative initializations for inserted nodes as an interesting direction for future work.

\subsection{Enhanced with LLMs (RQ5)}
Recently, the graph domain has become closely connected with LLMs on text-attributed graphs~\citep{tape,graphbridge,engine}, demonstrating significant improvements due to the advanced capabilities in language understanding. In this subsection, we explore the potential of combining \our with LLMs, like LLaMA2-7B~\citep{touvron2023llama}, on two text-attributed graphs, \ie, Arxiv and WikiCS~\citep{wikics}.

From Table~\ref{tab:llm}, we observe that \our consistently enhances downstream performance when combined with LLM features. The improved quality of node features enables \our to identify a better upsampled graph, thereby achieving superior performance. More complex combinations are left for future work.

\begin{table}[t]
\caption{The performance of UniGAP with LLM. $h_{\text{original}}$ indicates the use of features provided in PyG~\citep{pyg}, and $h_{\text{LLM}}$ denotes the use of hidden features generated by LLaMA2-7B.}
\label{LLM}
\begin{center}    
\renewcommand\arraystretch{1}
\resizebox{1.0\linewidth}{!}{
\begin{tabular}{lccccc}
\toprule
    & Feature type          & Arxiv                    & WikiCS\\
\midrule
GraphSAGE  & $h_{\text{original}}$            & 71.49 ± 0.3    & 79.13 ± 0.5      \\
\rowcolor{lightgray}
\quad + UniGAP & $h_{\text{original}}$  & 73.71 ± 0.2    & 80.57 ± 0.4      \\
\midrule
GraphSAGE  & $h_{\text{LLM}}$           & 75.03 ± 0.5        & 81.24 ± 0.7     \\
\rowcolor{lightgray}
\quad + UniGAP & $h_{\text{LLM}}$  & 76.21 ± 0.4        & 83.12 ± 0.5    \\
\bottomrule
\end{tabular}
}
\end{center}
\label{tab:llm}
\end{table}

\section{Conclusion}
In this work, we propose \our, a universal and adaptive graph upsampling methodology that alleviates the over-smoothing issue in the graph domain. \our enhances the adaptability, performance, and interpretability of GNNs. Furthermore, we integrate \our with LLMs, demonstrating the significant potential of combining LLMs with our method. 
Despite its notable performance and interpretability, there are areas where \our could be limited or further refined. 
First, although the additional time and space complexity introduced by UniGAP is moderate, applying \our to very large graphs remains challenging. In particular, storing the features of the newly inserted nodes incurs extra memory overhead, which can become substantial on massive graphs.
Second, while \our mitigates over-smoothing by slowing propagation through adaptive node insertion, the induced increase in path length may exacerbate over-squashing in graphs where long-range dependencies are critical. Understanding and balancing this trade-off in a more principled way is an important direction for future work.
Third, in this work we focus on node classification for studying over-smoothing. However, the \our framework can be extended to other tasks such as link-prediction and graph classification. Exploring these extensions is an interesting direction for future work.
In conclusion, we envision \our to inspire further exploration of graph upsampling methods.

\bibliographystyle{ACM-Reference-Format}
\bibliography{refs}

\newpage
\appendix

\section{Experimental Details}\label{appendixexp}
\subsection{Datasets}
We employs 8 node classification benchmark datasets, among which 4 are homophilic and 4 are heterophilic. The train-validation-test splits utilized are those which are publicly accessible. For measuring the degree of homophily of each dataset, we use the homophily metric suggested by \citep{homophilouseval}. Table~\ref{table3} delineates the comprehensive statistics of these benchmark collections.

The PubMed, CiteSeer, and Cora are citation networks~\citep{citionnetworks}. Within these networks, each vertex signifies a scholarly paper, and adjacency is established when there exists a bibliographic citation linking any two respective papers. The vertex attributes comprise the bag-of-words representation of the related manuscript, with the vertex categorization reflecting the scholarly domain of the associated paper.

The ogbn-arxiv~\citep{ogbbenchmark} dataset constitutes a directed graph typifying the citation network amid all Computer Science (CS) arXiv manuscripts as cataloged by the Microsoft Academic Graph (MAG). Each vertex within this graph is an arXiv document, while each directed edge signifies a citation from one document to another. Accompanying each paper is a 128-dimensional feature vector, derived from the mean of the embeddings corresponding to words in the document's title and abstract sections. 

The Texas, Cornell, and Wisconsin datasets~\citep{wikicsactor} originate from the WebKB corpus. In these, each vertex corresponds to a webpage, and adjacency is predicated upon a hyperlink’s presence interlinking any two respective webpages. Here, the vertex features encapsulate the bag-of-words attributes of the relevant webpage, with the vertex classification indicative of the webpage's categorical type.

The Actor dataset~\citep{wikicsactor} is a subgraph strictly encompassing actors from a larger network of film directors, actors, and screenwriters collated from Wikipedia entries. Each vertex in this dataset represents an actor, with an adjacency drawn between vertices when the pertinent actors are jointly mentioned on the same Wikipedia page. The nodal attributes are extrapolated from the salient keywords present on each actor's Wikipedia entry, and the vertex classification is ascertained by the contextual terms found on these pages.

\begin{table*}[t]
\caption{Details of datasets.}
\label{table3}
\begin{center}    
\renewcommand\arraystretch{1}
\begin{tabular}{lcccccc}
\hline
Dataset        & Homopily     & \#Nodes         & \#Edges    & \#Features   &\#Classes &Split(\%)\\
\hline
Pubmed      &0.66 &19,717 &44,324 &500  &3 &0.3/2.5/5.0\\
Citeseer    &0.63 &3,327   &4,552   &3,703 &6 &5.2/18/37\\
Cora        &0.77 &2,708   &5,728   &1,433 &7 &3.6/15/30\\
Arxiv       &0.42 &169,343&1,166,243 &128 &40 &53.7/17.6/28.7\\
WikiCS      &0.57 &11,701 &216,123 &300   &10 &5.0/15/50\\
\hline
Texas       &0.00 &183    &279    &1,703 &5 &48/32/20\\
Cornell     &0.02 &183    &277    &1,703 &5 &48/32/20\\
Wisconsin   &0.05 &251    &450    &1,703 &5 &48/32/20\\
Actor       &0.01 &7,600  &26,659  &932  &5 &48/32/20\\
Squirrel    &0.03 &5,201  &198,493 &2,089  &5 &60/20/20\\
Questions   &0.02 &48,921 &153,540 &301    &2 &60/20/20\\
Amazon-ratings &0.14 &24,492 &93,050 &300  &5 &60/20/20\\
\hline
\end{tabular}
\end{center}
\end{table*}

\subsection{Hyperparameters}
Experimental results are reported on the hyperparameter settings below, where we choose the settings that achieve the highest performance on the validation set. We choose hyperparameter grids that do not necessarily give optimal performance, but hopefully cover enough regimes so that each model is reasonably evaluated on each dataset. 
\begin{itemize}
    \item lr $\in \{5e-2,1e-2,5e-3,1e-3,5e-4\}$
    \item hidden\_dim $\in \{16,32,64,128\}$
    \item dropout $\in \{0,0.1,0.2,0.3,0.5,0.8\}$
    \item weight\_decay $\in \{1e-2,5e-3,1e-3,5e-4,1e-4\}$
\end{itemize}

For GCNII, 
\begin{itemize}
    \item alpha $\in \{0.1,0.2,0.3,0.4,0.5\}$
    \item theta $=1$
\end{itemize}

For UniGAP, 
\begin{itemize}
    \item MVC\_dim $\in \{32,64,128,256\}$
    \item $w$ $\in \{1,2,3,4,\text{None}\}$
\end{itemize}

For Half-Hop, 
\begin{itemize}
    \item p $\in \{0,0.5,0.75,1\}$
    \item $\alpha=0.5$
\end{itemize}

\subsection{Parameter study}
\label{appendixbeta}
\begin{figure}[t] 
\centering 
\includegraphics[width=1\linewidth]{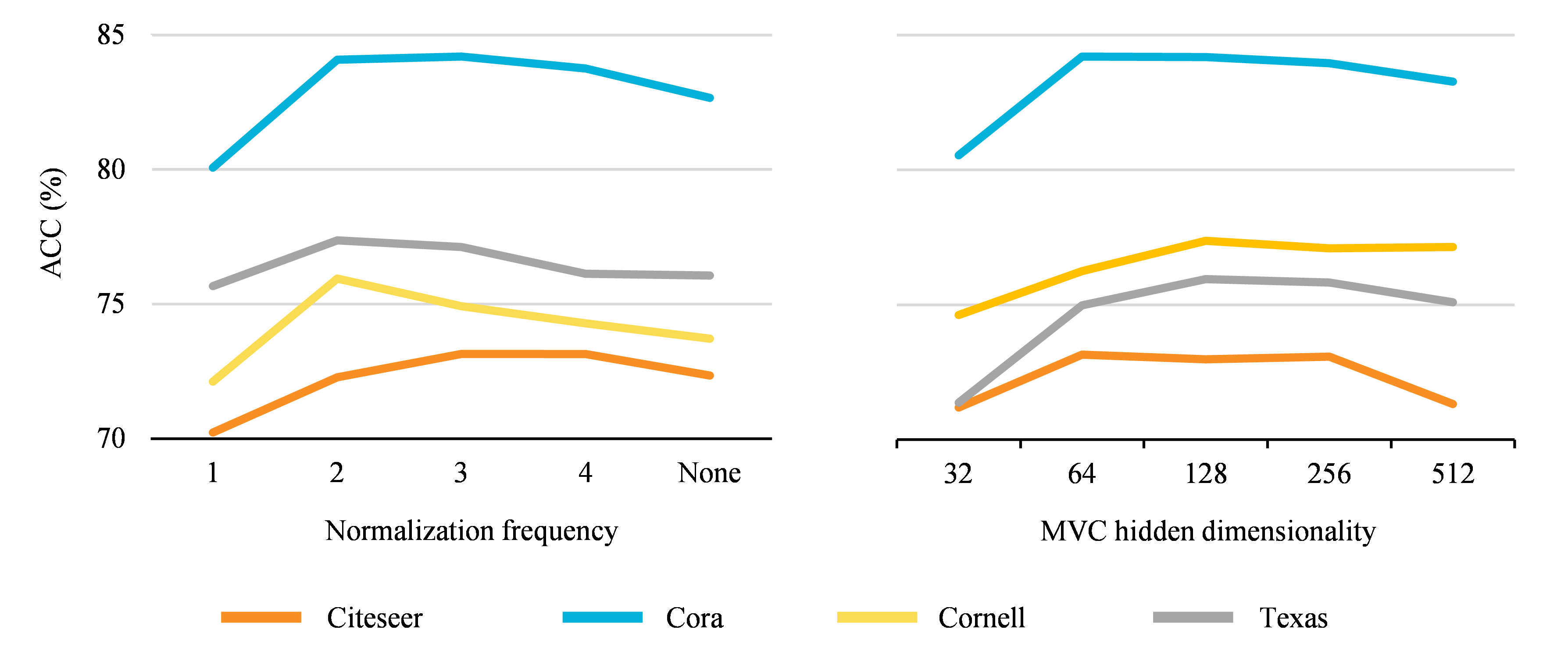} 
\caption{The performance and MAD of UniGAP+GCN of different layers with different choice of $\beta$ in Citeseer and Texas.}
\label{parameter}
\end{figure}

\noindent\textbf{The normalization frequency of the trajectory.}
Recall that in Eq.~(1) we apply an $L_2$ normalization to the layer-wise representations $H^{(l)}$ every $w$ layers to prevent feature magnitudes from exploding during trajectory computation. The hyperparameter $w$ therefore controls the normalization frequency of the trajectories. 
We conduct a sensitivity study by varying $w$ on representative homophilic and heterophilic datasets. 
As shown in Figure~\ref{parameter}, very small values ($w = 1$) over-normalize the representations and slightly harm performance, while very large values (or no normalization) lead to unstable trajectories and suboptimal performance. In contrast, a moderate range of $w$ (2 or 3) yields stable trajectories and robust accuracy, and the default setting used in our main experiments lies in this stable regime.

\noindent\textbf{The hidden dimensionality in MVC Encoder.}
We further study the effect of the hidden dimensionality in the MVC encoder, which controls the capacity of the trajectory aggregation module. 
Concretely, we vary the hidden size of MVC report the corresponding UniGAP performance. 
As shown in Figure~\ref{parameter}, using an extremely small hidden size creates a strong information bottleneck and leads to noticeable accuracy drops, especially on heterophilic graphs where more complex trajectory patterns need to be captured. 
Increasing the hidden dimensionality from a small to a moderate value consistently improves performance, but further enlarging the MVC encoder beyond the default size brings only marginal gains while increasing computational and memory cost. 

\noindent\textbf{The weight of smoothing loss.}
$\beta$ is the regularization coefficient that controls the impact of the smoothing metric. A larger $\beta$ means that we put more emphasis on solving the over-smoothing problem for downstream tasks, even if the performance is suboptimal in this setting. Here we provide an ablation experiment to explore the effect of different $\beta$ values on model performance and smoothness metrics. 

\begin{figure}[t] 
\centering 
\includegraphics[width=1\linewidth]{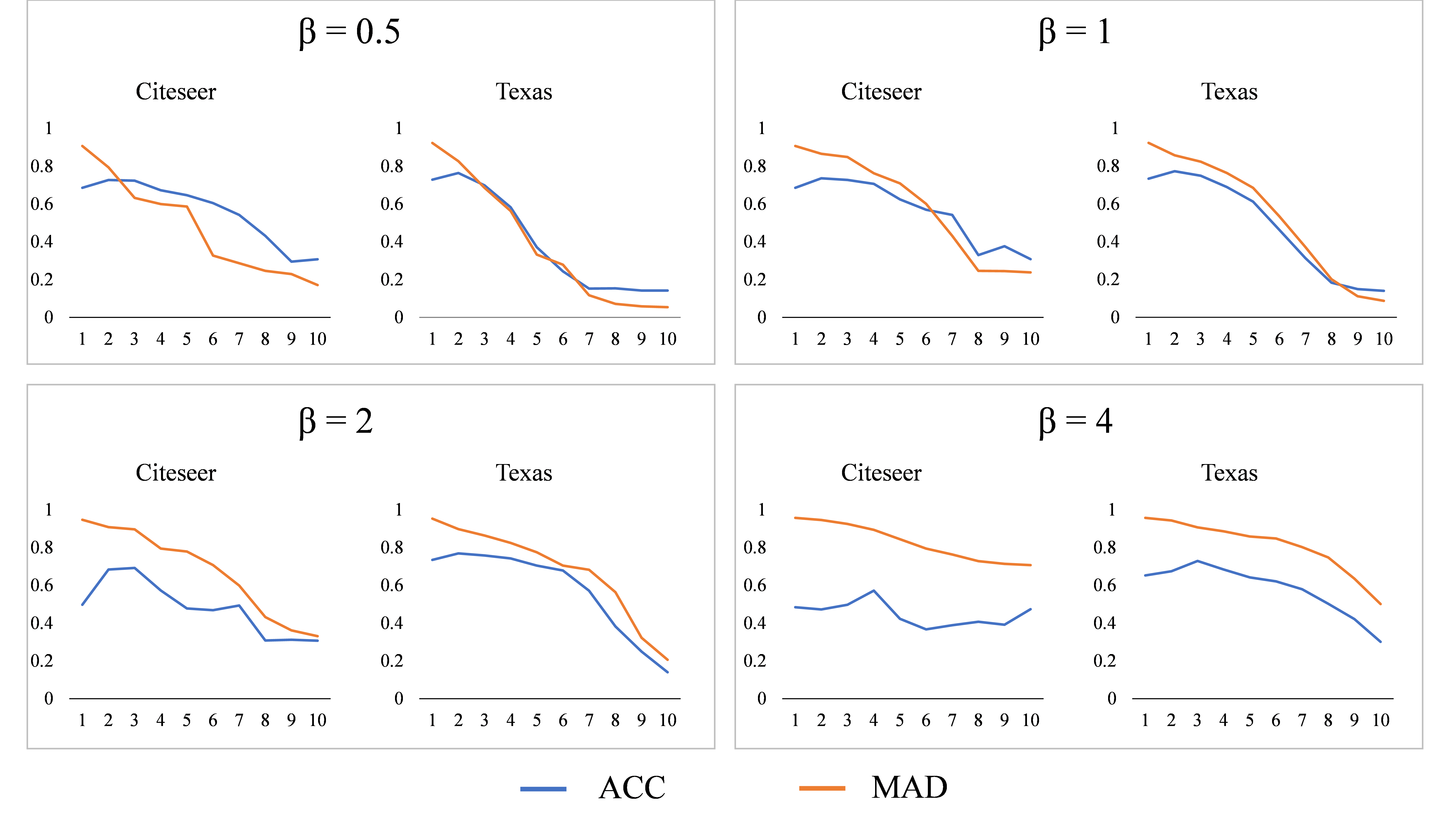} 
\caption{The performance and MAD of UniGAP+GCN of different layers with different choice of $\beta$ in Citeseer and Texas.}
\label{beta}
\end{figure}

As shown in Figure~\ref{beta}, as the selected $\beta$ increases, the training process tends to maintain a lower degree of over-smoothing (indicated by a larger MAD), but this may compromise optimization for task accuracy, leading to reduced performance. Additionally, the smoothness metrics and accuracy are more consistent in heterophilic graphs compared to homophilic graphs, which suggests that higher $\beta$ values can be chosen for heterophilic graphs in practical applications. Moreover, as indicated in Table~\ref{withoutbeta}, even with $\beta$ set to 0—meaning no additional optimization for over-smoothing is applied—UniGAP still significantly improves model performance.

\begin{table}[t]
\caption{The performance of UniGAP without MAD loss.}
\label{withoutbeta}
\begin{center}    
\renewcommand\arraystretch{1}
\begin{tabular}{lcc}
\toprule
        & Citeseer              & Texas\\
\midrule
GCN   & 72.10    & 65.65      \\
\quad + HalfHop  & 72.40        & 71.62    \\
\rowcolor{lightgray}
\quad + UniGAP  & 73.15        & 77.37    \\
\midrule
\quad + HalfHop w/o MAD\_loss  & 72.36        & 72.04    \\
\rowcolor{lightgray}
\quad + UniGAP w/o MAD\_loss  & 72.80        & 77.18    \\

\bottomrule
\end{tabular}
\end{center}
\vspace{-1em}
\end{table}

\subsection{\our as a Plug-in to Non-upsampling Methods}
Since \our is designed as a plug-in module, it is conceptually orthogonal to methods that modify the graph structure or propagation scheme without upsampling. To demonstrate this complementarity, we further combine \our with two recent non–upsampling approaches for alleviating over-smoothing and over-squashing: Delaunay-based rewiring (DR)~\citep{dr} and Multi-Track Message Passing (MT)~\citep{mtgcn}. As shown in Table~\ref{asplugin}, both DR and MT substantially improve the performance of a vanilla GCN, especially on the heterophilic Texas dataset (from 65.65\% to around 85\%). When we additionally equip these methods with \our, the performance is further boosted on both Citeseer and Texas (e.g., from 73.35\% to 73.82\% on Citeseer with DR, and from 85.07\% to 86.89\% on Texas). These results indicate that \our provides complementary benefits and can be seamlessly combined with non–upsampling methods, reinforcing its role as a general plug-in for enhancing different families of GNN architectures.

\begin{table}[t]
\caption{Performance of UniGAP combined with non–upsampling methods for alleviating over-smoothing. DR denotes Delaunay-based rewiring and MT denotes Multi-Track Message Passing.}
\label{asplugin}
\begin{center}
\renewcommand\arraystretch{1}
\begin{tabular}{lcc}
\toprule
        & Citeseer   & Texas \\
\midrule
GCN                 & 72.10 & 65.65 \\
\quad + DR          & 73.35 & 85.07 \\
\rowcolor{lightgray}
\quad + DR + \our   & 73.82 & 86.89 \\
\midrule
\quad + MT          & 72.74 & 85.16 \\
\rowcolor{lightgray}
\quad + MT + \our   & 73.89 & 85.97 \\
\bottomrule
\end{tabular}
\end{center}
\vspace{-1em}
\end{table}

\subsection{Training stability and scalability}
\label{training}
\begin{figure}[t] 
\centering 
\includegraphics[width=1\linewidth]{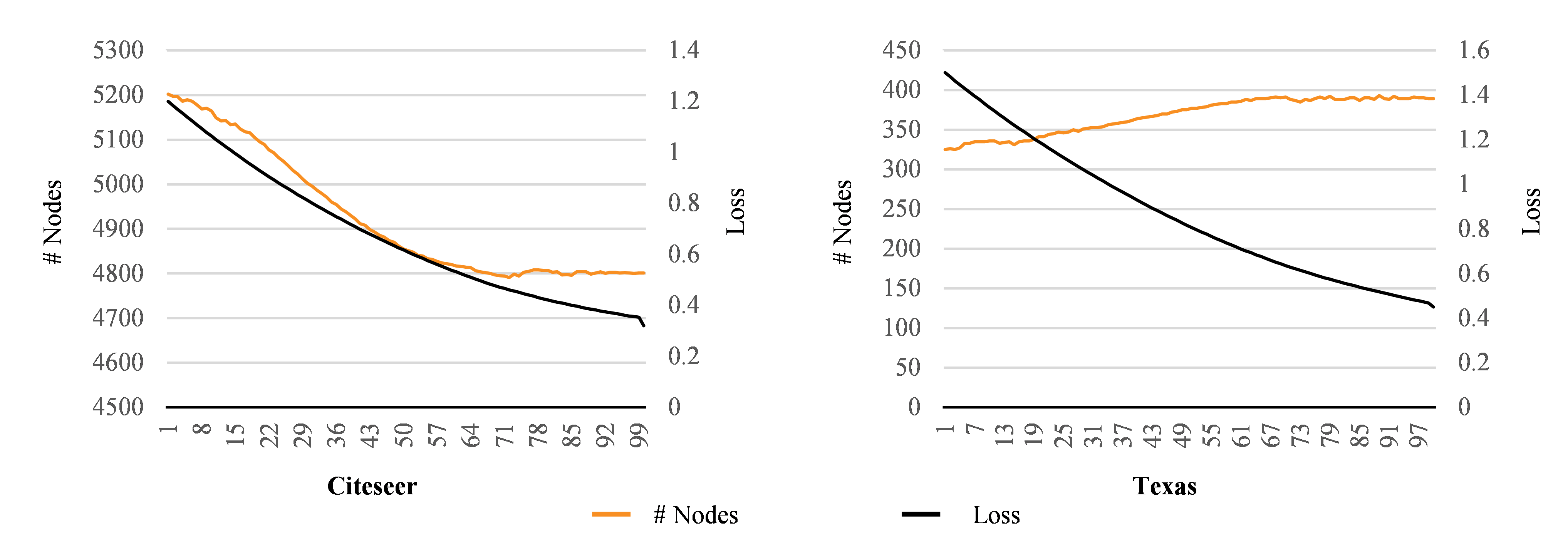} 
\caption{Learning dynamics of the loss and the total number of nodes.}
\label{dynamics} 
\end{figure}

In our experiments, we did not observe oscillatory behavior or divergence. The upsampling masks and the resulting number of inserted nodes stabilize after a moderate number of epochs, and the validation accuracy curves remain smooth across random seeds. Here, we include training dynamics showing the number of nodes across epochs. As shown in Figure~\ref{dynamics}, the loss decreases smoothly over time, and the total number of nodes plateaus after a certain number of epochs, indicating that the learned graph structure converges rather than fluctuating erratically.

As for the scalability of \our, as detailed in Section~4.3, the complexity of \our is on par with traditional GNN methods, maintaining efficiency. Here, we present the running time of UniGAP on the Arxiv dataset. As shown in Table~\ref{timecomplexity}, UniGAP has outstanding performance while maintaining competitive running times.

\begin{table}[t]
\caption{The running time of GCN with different upsampling method on the Arxiv dataset.}
\label{timecomplexity}
\begin{center}    
\renewcommand\arraystretch{1}
\resizebox{1.0\linewidth}{!}{
\begin{tabular}{llcc}
\toprule
        & Time Complexity              & Running time(epoch)	& ACC\\
\midrule
GCN   & $\mathcal{O}(LEd+LNd^2)$    & 5.64s  & 71.74   \\
\quad + AdaEdge &$O(N^{2}+LEd+LNd^2)$  & 16.73s        & 71.98    \\
\quad + HalfHop  &$O(LEd+LNd^2)$ & 7.97s        & 72.14    \\
\rowcolor{lightgray}
\quad + UniGAP  &$O(LN+LEd+LNd^2)$ & 12.28s        & 73.02    \\
\bottomrule
\end{tabular}
}
\end{center}
\vspace{-1em}
\end{table}

\section{Theoretical Proofs}\label{appendixproof}
Our proofs are based on the work of ~\citep{theoreticoversmoothing} and rely heavily on the Lemmas and proofs in the original paper. For a better understanding of our proof, we appeal interested readers to read the work of ~\citep{theoreticoversmoothing}. For clarity, we first summarize the main symbols used in Section~\ref{sec:justification} and the following lemmas and theorems in Table~\ref{symbols}.

\begin{table}[t]
\caption{The symbol dictionary.}
\centering
\label{symbols}
\renewcommand\arraystretch{1}
\begin{tabular}{p{0.17\linewidth} p{0.78\linewidth}}
\toprule
Symbol & Meaning \\
\midrule
$H^{(k)}_{e}$ & Node features after $k$ message-passing steps at training epoch $e$ in the linearized GNN. \\
$f^{k}(A,\theta_{u},X)$ & Shorthand for the $k$-hop node features $H^{(k)}_{e}$, viewed as a function of the base adjacency $A$, UniGAP parameters $\theta_{u}$, and input features $X$ (Eq.~\ref{eq:node_feat}). \\
$\theta$ & Parameters of the linear predictor (classifier/regressor) on top of $H^{(k)}_{e}$ in the theoretical analysis. \\
$\theta_{u}$ & Parameters of UniGAP (upsampler) that determine the adaptive message-passing matrix $\hat{A}_{e}$. \\
$Y$ & Ground-truth labels of the $n$ training nodes (stacked into a vector or matrix). \\
$n$ & Number of labeled training nodes used in the empirical risk. \\
$\lambda$ & $\ell_{2}$ regularization coefficient for the predictor parameters $\theta$. \\
$\gamma$ & $\ell_{2}$ regularization coefficient for the UniGAP parameters $\theta_{u}$ in the theoretical objective (Eq.~\ref{eq:pro}). \\
$\beta$ & Weight of the MAD-based smoothing regularizer in the joint objective, controlling the strength of explicit over-smoothing penalization. \\
\midrule
$\Sigma$ & Latent covariance matrix of node features in the underlying statistical model. \\
$\beta^{*}$ & True regression coefficients in the latent model, used to define the population risk (Definition~1). \\
$M$ & Projection matrix that maps latent features to the observed feature space in the theoretical model. (This $M$ is distinct from the upsampling mask in Sec.~\ref{sec:upsampler}, and is only used in the analysis.) \\
$S$ & Generic symmetric positive semi-definite matrix (typically covariance-like) on which the risk $\mathcal{R}_{\text{reg.}}(S)$ is evaluated. \\
$\mathcal{R}_{\text{reg.}}(S)$ & Regularized risk associated with covariance $S$ (Definition~1), measuring prediction error under ridge regression with penalty $\gamma$. \\
$\mathcal{R}^{(k)}$ & Risk after $k$ rounds of message passing \emph{without} UniGAP (Lemma~\ref{lemma3.3}). \\
$\Sigma^{(k)}_{\text{UniGAP}}$ & Approximate covariance of node features after $k$ rounds of message passing \emph{with} UniGAP (Lemma~\ref{lemma3.4}). \\
$A$ & Effective smoothing operator defined as $(I + \Sigma^{-1})^{-1}$ in Lemma~\ref{lemma3.3}, characterizing the contraction induced by message passing. \\
$p$ & Average upsampling probability (expected value of $\hat{P}[:,0]$ at the optimum), reflecting how often edges are upsampled by UniGAP in the theoretical analysis. \\
$I$ & Identity matrix of appropriate dimension. \\
\bottomrule
\end{tabular}
\end{table}

\subsection{Proof of Proposition 4.1}
\begin{proof}
Here we consider a simplified situation of linear GNN with mean aggregation, the smoothed node features after k rounds of mean aggregation are:

\begin{equation}
    H^{(k)} =L^{(k)}X
\end{equation}

We consider learning with a single Mean Square Error (MSE) loss and Ridge regularization. For $\lambda > 0$, the regression coefficients vector on the smoothed features is

\begin{equation}
    \begin{aligned}
        \hat{\theta} &\overset{def_{.}}{=} \argmin_{\theta} \frac{1}{2n}||Y-H^{(k)}\theta||^{2}+\lambda||\theta||^{2} \\ &= \left( \frac{(H^{(k)})^{T}H^{(k)}}{n}+\lambda I \right)^{-1}\frac{(H^{(k)})^{T}Y}{n}
    \end{aligned}
\end{equation}

$n$ and $n_{test}$ is the count of training data and test data, and $\theta$ is the set of learnable parameters of the training model. Then, the test risk $R^{(k)}$ is 

\begin{equation}
    R^{(k)} \overset{def_{.}}{=} \frac{1}{n_{test}} ||Y_{test}-\hat{Y}^{(k)}||^{2} \ where\ \hat{Y}^{(k)}= H^{(k)} \hat{\theta}
\end{equation}

As proofed in \citep{theoreticoversmoothing}, there is an optimal $k^{*} > 0$ such that $R^{(k^{*})} < min(R^{(0)}, R^{(\infty)})$. Below, we show that when equip with UniGAP, the model can learn the optimal $\langle\hat{\theta}, \hat{\theta_u}\rangle$ for each k, adaptively changing the node's receptive field and message propagation process so that the model operates on the optimal graph structure. Thus by learning graph structure, UniGAP makes $k^{*}$ approximate our choice of k, rather than heuristically finding $k^{*}$ for each downstream task.

When equpied with UniGAP, the optimization objective is:

\begin{equation}
\begin{aligned}
 \langle\hat{\theta}, \hat{\theta_u}\rangle_{e} =& \argmin_{\langle\theta, \theta_u\rangle} \bigg( \frac{1}{2n} \| Y - H^{(k)}_e \theta \|^2 + \lambda \|\theta\|^2 + \gamma \|\theta_u\|^2  \\& - \beta \text{MAD}(H^{(k)}_e) \bigg) 
\end{aligned}
\end{equation}

First, we ignore the $\beta \text{MAD}(H^{(k)}_e)$ term as it may introduce complex non-linear dependencies. We only consider the preceding terms and the problem can be reformulated as:

\begin{equation}
\langle\hat{\theta}, \hat{\theta_u}\rangle = \argmin_{\langle\theta, \theta_u\rangle} \left( \frac{1}{2n} \| Y - H^{(k)}(\theta_u) \theta \|^2 + \lambda \|\theta\|^2 + \gamma \|\theta_u\|^2 \right) 
\end{equation}

Next, we derive each component and set the derivatives to zero to find the minima. When solving for $\theta$, we assume $H^{(k)}_e(\theta_u)$ is known. First, we write the Lagrangian form:

\begin{equation}
L(\theta, \theta_u) = \frac{1}{2n} \| Y - H^{(k)}_e \theta \|^2 + \lambda \|\theta\|^2 + \gamma \|\theta_u\|^2 
\end{equation}

Then we take the derivative with respect to $\theta$ and set it to zero:

\begin{equation}
\nabla_{\theta} L = -\frac{1}{n} (H^{(k)}_e)^T (Y - H^{(k)}_e \theta) + 2\lambda \theta  = 0
\end{equation}

We apply a rearranging and obtain:

\begin{equation}
 \frac{1}{n} (H^{(k)}_e)^T Y = \left( \frac{1}{n} (H^{(k)}_e)^T H^{(k)}_e + \lambda I \right) \theta 
\end{equation}

So we can get:

\begin{equation}
\theta = \left( \frac{1}{n}(H^{(k)}_e)^T H^{(k)}_e + \lambda I \right)^{-1} \frac{1}{n}(H^{(k)}_e)^T Y 
\end{equation}

When solving for $\theta_u$, for a fixed solution \(\hat{\theta}\), we take the derivative with respect to \(\theta_u\) and set it to zero:

\begin{equation}
\begin{aligned}
\nabla_{\theta_u} L =& \frac{\partial}{\partial \theta_u }\left( \frac{1}{2n} \| Y - H^{(k)}(\theta_u) \hat{\theta} \|^2 \right) + \nabla_{\theta_u}(\gamma \|\theta_u\|^2) \\&- \nabla_{\theta_u}(\beta \text{MAD}(H^{(k)}(\theta_u))) 
\end{aligned}
\end{equation}

Setting the derivative to zero, we can get:

\begin{equation}
\begin{aligned}
\frac{1}{n} \left( \left( \frac{\partial H^{(k)}(\theta_u)}{\partial \theta_u} \right)^T (Y - H^{(k)}(\theta_u) \hat{\theta}) \hat{\theta} \right) - 2 \gamma \theta_u \\+ \beta \frac{\partial}{\partial \theta_u} \text{MAD}(H^{(k)}(\theta_u)) = 0 
\end{aligned}
\end{equation}

We apply a rearranging and obtain:

\begin{equation}
\begin{aligned}
\theta_u =& \gamma^{-1} \bigg( \frac{1}{2n} \left( \frac{\partial H^{(k)}(\theta_u)}{\partial \theta_u} \right) ^T (Y - H^{(k)}(\theta_u) \hat{\theta}) \hat{\theta} \\&- \frac{\beta}{2} \frac{\partial}{\partial \theta_u} \text{MAD}(H^{(k)}(\theta_u)) \bigg)
\end{aligned}
\end{equation}
\end{proof}

\subsection{Proof of Lemma 4.3}
\begin{proof}
Following the Theorem 4 and Section 4.2 of the work of \citep{theoreticoversmoothing}, they adopt the latent space random graph model, and assume that the observed node features are projections of some underlying latent features with Gaussian distribution $\mathcal{N}(0,\Sigma)$. They get the regression risk with $k$ step of smoothing as:
$$
\mathcal{R}^{(k)} \simeq \mathcal{R}_{reg.}(\Sigma^{(k)}) 
=  \mathcal{R}_{reg.}((I + \Sigma^{-1})^{-2k}\Sigma)
$$
Here $A = (I + \Sigma^{-1})^{-1}$, then the risk become $ \mathcal{R}^{(k)} \simeq \mathcal{R}_{reg.}(A^{2k}\Sigma)$. Thus, the rate of smoothing of original MP
is $2k$ due to the $A^{2k}\Sigma$ operator.

Now following the \citep{azabou2023halfhop}, we compute the regression risk with $k$ step of smoothing with UniGAP. We define $p = \mathbb{E}(\hat{P}[:,0])$ and $\hat{P}$ denotes the upsampling probability at the optimal graph structure, $U_{i}$ is the inserted nodes between node $i$ and $V_{i}$. After we insert nodes in original edges, begin with the first round of message passing, we can get:
\begin{equation}
    \begin{aligned}
        \forall i, x_{i}^{(1)} &= \sum_{u \in U_{i}} a_{ij}\hat{x}^{(0)}_{u} \\
        \mathbb{E}(x_{i}^{(1)}) 
        &= \sum_{j \in V_{i}} a_{ij}((1-p)x_{i}^{(0)}+px_{j}^{(0)})\\
        &= (1-p)x_{i}^{(0)}(\sum_{j \in V_{i}} a_{ij})+p\sum_{j \in V_{i}} a_{ij}x_{j}^{(0)}\\
        &= (1-p)x_{i}^{(0)}+p\sum_{j \in V_{i}} a_{ij}x_{j}^{(0)}
    \end{aligned}
\end{equation}

Here $U_{i}$ is the upsampled inserted nodes between node $i$ and its neighbor $j$. For inserted nodes, their features are:
\begin{equation}
\forall i, j \in V_{i}, u \in U_{i}, \hat{x}^{(1)}_{u} = x^{(0)}_{j}
\end{equation}

At the second round of message passing, the original nodes are updated as follows:
\begin{equation}
    \begin{aligned}
        \forall i, j \in V_{i}, u \in U_{i}, x_{i}^{(2)} &= \sum_{u \in U_{i}} a_{ij}\hat{x}^{(1)}_{u} = \sum_{u \in U_{i}} a_{ij}x^{(0)}_{j} \simeq Ax_{i}^{(0)}
    \end{aligned}
\end{equation}

The the features of inserted nodes are same as the the source node:
\begin{equation}
\forall i, j \in V_{i}, u \in U_{i}, \hat{x}^{(2)}_{u} = x^{(1)}_{j} \simeq (1-p)x_{j} + pAx_{j}
\end{equation}

Repeat the process of message passing of the first round and the second round, we can get a recurrence relation:
\begin{equation}
\begin{aligned}  
     x_{i}^{(k+1)}& \simeq A\hat{x}^{(k)}_{u}\\
     \hat{x}^{(k+1)}_{u} &= x_{j}^{(k)}\\
\end{aligned}
\end{equation}
By associating the above equations we can get the connection of $i$ and $j$, as $x_{i}^{(k+2)} \simeq A\hat{x}^{(k+1)}_{u} = Ax_{j}^{(k)}$. Then we use $2k+1$ to denote an odd number and get
\begin{equation}
    \begin{aligned}  
x_{i}^{(2k+1)} &\simeq A^{k}{x}^{(1)}_{j}\\
     &=A^{k}((1-p)x_{i}^{(0)} + pAx_{i}^{(0)})\\
     &=(1-p)A^{k}x_{i}+pA^{k+1}x_{i}
     \end{aligned}  
\end{equation}

Then we replace $A$ as $(I + \Sigma^{-1})^{-1}$ and get
\begin{equation}
\Sigma_{UniGAP}^{(k)} = \frac{1}{2}A^{k-1}(I+((1-p)I+pA)^{2})\Sigma
\end{equation}
\end{proof}

\section{Algorithm}
This section we show the pseudo-code identity of the UniGAP in algorithm~\ref{alg_our_method}.

\begin{algorithm}
\caption{The UniGAP Algorithm}
\label{alg_our_method}
\KwIn{Input graph $G$, MVC encoder parameters $\omega$, UniGAP module parameters $\phi$, downstream model parameters $\psi$, number of epochs $num\_epochs$}
\KwOut{Optimized model}

\SetKwFunction{FMain}{Main}
\SetKwFunction{FTrajectory}{TrajectoryPrecomputation}
\SetKwFunction{FEncode}{MVCEncoder}
\SetKwFunction{FUgap}{AdaptiveUpsampler}
\SetKwFunction{FDownstream}{DownstreamModel}

\SetKwProg{Fn}{Function}{:}{}

\Fn{\FTrajectory{$G$}}{
    $T \gets \mathcal{P}(G)$ \tcp*{Compute initial trajectories}
    \KwRet $T$
}

\Fn{\FEncode{$T, \omega$}}{
    $\hat{T} \gets \mathcal{F}_{\omega}(T)$ \tcp*{Compute Multi-View Condensation}
    \KwRet $\hat{T}$
}

\Fn{\FUgap{$\hat{T}, \phi$}}{
    $P \gets \mathcal{U}_{\phi}(\hat{T})$ \tcp*{Compute upsampling probabilities}
    $\hat{G} \gets \mathcal{S}(P)$ \tcp*{Upsampling with gumbel-softmax}
    \KwRet $\hat{G}$
}

\Fn{\FDownstream{$\hat{G}, \psi$}}{
    \For{$i = 1$ \KwTo $num\_layers$}
    {
        $t_{i} \gets Layer_{i}(\hat{G},\psi_{i})$ \;
    }
    $T \gets \operatorname{concat}(t_{1},t_{2}, \cdots t_{l}) $ \;
    $\mathcal{L} \gets \operatorname{compute\_loss}(\hat{G},\omega, \phi, \psi)$ \;
    optimize\_modules($\mathcal{L}$) \;
    \KwRet $T,\mathcal{L}$
}

\Fn{\FMain{$G, \omega, \phi, \psi$}}{
    $T \gets$ \FTrajectory{$G$} \;

    \For{$epoch = 0$ \KwTo $num\_epochs$}{
        $\hat{T} \gets$ \FEncode{$T,  \omega$} \;
        $\hat{G} \gets$ \FUgap{$\hat{T}, \phi$} \;
        $T$,$\mathcal{L} \gets$ \FDownstream{$\hat{G}, \psi$} \;
    }
    
    \KwRet $optimized\_model$ \;
}

$optimized\_model \gets$ \FMain{$G, \omega, \phi, \psi$} \;
\end{algorithm}

\end{document}